\documentclass[final,3p,times,twocolumn]{elsarticle}
\pdfoutput=1 
\UseRawInputEncoding
\usepackage{lineno}

\usepackage[hyphens]{url}
\usepackage{breakurl}
\usepackage[colorlinks,
linkcolor=black,
anchorcolor=blue,
citecolor=black,
urlcolor=black]{hyperref}

\modulolinenumbers[5]
\journal{Information Fusion}
\usepackage[ruled,linesnumbered]{algorithm2e}
\usepackage[table]{xcolor}
\usepackage{amssymb}
\usepackage{graphics}
\usepackage{graphicx}
\usepackage{epsfig}
\usepackage{multirow}

\newtheorem{definition}{Definition}
\usepackage{amsmath}
\usepackage{bbm}

\usepackage{array}
\usepackage{mdwmath}
\usepackage{mdwtab}
\usepackage{eqparbox}
\usepackage{url}
\usepackage{amsmath}
\usepackage{dutchcal}
\usepackage{amsfonts,amssymb} 
\usepackage{multirow}
\usepackage{booktabs}
\usepackage{algorithmic}  
\usepackage{algorithm2e}
\usepackage{graphicx}
\usepackage{threeparttable}
\usepackage{ulem}
\usepackage{subfig}
\usepackage{multirow} 
\usepackage{makecell} 
\usepackage{tabularx} 
\usepackage{svg}
\usepackage{pifont}
\usepackage{soul}
\usepackage{color,xcolor}
\soulregister{\cite}7
\soulregister{\ref}7
\soulregister{\subsection}7
\soulregister{\tableautorefname}7
\soulregister{\section}7
\soulregister{\textit}7
\soulregister{\eqref}7

\usepackage{ragged2e} 
\usepackage{booktabs} 
\usepackage{svg}
\usepackage{CJKutf8}
\usepackage{array}
\usepackage{makecell}
\biboptions{numbers,sort&compress}

\begin{document}

\begin{frontmatter}

\title{A Survey of Route Recommendations: Methods, Applications, and Opportunities}

\author[1,2,3,4]{Shiming Zhang}
\ead{zhangshiming@my.swjtu.edu.cn}

\author[1,2,3,4]{Zhipeng Luo\corref{Corresponding_author}}
\cortext[Corresponding_author]{Corresponding author}
\ead{zpluo@swjtu.edu.cn}

\author[1,2,3,4]{Li Yang}
\ead{yangli_ef@my.swjtu.edu.cn}

\author[1,2,3,4]{Fei Teng}
\ead{fteng@swjtu.edu.cn}

\author[1,2,3,4]{Tianrui~Li}
\ead{trli@swjtu.edu.cn}

\address[1]{School of Computing and Artificial Intelligence, Southwest Jiaotong University}
\address[2]{Engineering Research Center of Sustainable Urban Intelligent Transportation, Ministry of Education}
\address[3]{National Engineering Laboratory of Integrated Transportation Big Data Application Technology, Southwest Jiaotong University}
\address[4]{Manufacturing Industry Chains Collaboration and Information Support Technology Key Laboratory of Sichuan Province, Southwest Jiaotong University, Chengdu 611756, Sichuan, P.R. China}

\begin{abstract}
Nowadays, with advanced information technologies deployed citywide, large data volumes and powerful computational resources are intelligentizing modern city development. As an important part of intelligent transportation, route recommendation and its applications are widely used, directly influencing citizens’ travel habits. Developing smart and efficient travel routes based on big data (possibly multi-modal) has become a central challenge in route recommendation research. 
Our survey offers a comprehensive review of route recommendation work based on urban computing. It is organized by the following three parts: 1) Methodology-wise. We categorize a large volume of classic methods and modern deep learning methods. Also, we discuss their historical relations and reveal the edge-cutting progress.
2) Application-wise. We present numerous novel applications related to route commendation within urban computing scenarios. 3) We discuss current problems and challenges and envision several promising research directions. We believe that this survey can help relevant researchers quickly familiarize themselves with the current state of route recommendation research and then direct them to future research trends.

\end{abstract}


\end{frontmatter}


\section{Introduction}
\label{introduction}

The world has witnessed a vigorous development of information technologies such as electronic devices, data storage, and the 5G Internet, which enable various information systems to record vast spatio-temporal data regarding citizens’ trajectories~\cite{zheng2009mining}. These data reflect people’s activities such as daily commutes, travels, and gatherings. Meanwhile, people living in the big data era also demand for more intelligent route recommendations when facing richer travel options~\cite{zheng2015trajectory}. How to design smart, efficient, and personalized travel routes with provided navigation services is of great research and practical value for developing modern travel schemes.

\begin{table*}[h]\small
\renewcommand
\arraystretch{1.0}
\centering
\caption{A summary of survey articles related to route recommendation.}
\small
\begin{tabularx}{\textwidth}{Xlc}
\toprule[2pt]

\textbf{Title}                                                                                      &\textbf{Category}                                                 &\textbf{Reference} \\
\midrule
\multirow{2}{*}{A Survey of Traffic Data Visualization}                                  & Data Visualization;             &        \multirow{2}{*}{\cite{chenSurveyTrafficData2015a}}     \\ 
& Visual Analysis& \\ \cline{1-3}

\multirow{2}{*}{A Survey on Trajectory Data Mining: Techniques and Applications}                            & Trajectory Data;                 &        \multirow{2}{*}{\cite{fengSurveyTrajectoryData2016}}      \\ 
& Applications Analysis& \\ \cline{1-3}

Beyond the Shortest Route: A Survey on Quality-Aware Route        & Route Navigation;             &     \multirow{2}{*}{\cite{siriarayaShortestRouteSurvey2020}}         \\ 
Navigation for Pedestrians& Applications Scenario& \\ \cline{1-3}

\multirow{2}{*}{Deep Learning for Spatio-Temporal Data Mining: A Survey}                                    & Spatio-Temporal Data;           &     \multirow{2}{*}{\cite{wangDeepLearningspatiotemporal2022}}         \\ 
& Applications Analysis& \\ \cline{1-3}

Designing Mobile Application Messages To Impact Route Choice: &Mobile Messages;          &   \multirow{2}{*}{\cite{mayrDesigningMobileApplication2023} }          \\ 
 A Survey And Simulation Study& Impact Route Choice& \\ \cline{1-3}

\multirow{2}{*}{Mobile Recommender Systems in Tourism}                                                      & Mobile Tourism;  &    \multirow{2}{*}{\cite{gavalasMobileRecommenderSystems2014} }         \\ 
&  Tour Planning& \\ \cline{1-3}

\multirow{2}{*}{Online Delivery Route Recommendation in Spatial Crowdsourcing}                              & Route Recommendation;   &   \multirow{2}{*}{\cite{sunOnlineDeliveryRoute2019b}}          \\ 
& Spatial Crowdsourcing& \\ \cline{1-3}

\multirow{2}{*}{Route Guidance and Information Systems}                              &  Route Navigation;                &    \multirow{2}{*}{ \cite{ben-akivaRouteGuidanceInformation2001} }        \\ 
& Driver Behaviour Models & \\ \cline{1-3}

\multirow{2}{*}{Route Search and Planning: A Survey }                                                       &  Route Matching;                                                          & \multirow{2}{*}{\cite{liRouteSearchPlanning2021}  }     \\ 
& Route Planning& \\ \cline{1-3}

Stochastic Dynamic Vehicle Routing in The Light of Prescriptive       &  Vehicle Routing;   &        \multirow{2}{*}{\cite{soeffkerStochasticDynamicVehicle2022} }    \\
Analytics: A Review& Prescriptive Analytics & \\ 

\bottomrule[2pt]

\end{tabularx}
\label{tab:related review}
\end{table*}


Route recommendation or planning is key to many route-based applications such as navigation, deliveries, and travel planning~\cite{liuIntegratingDijkstraAlgorithm2020}. The main goal is to design routes that can be realized on a given transportation network with certain travel requirements. This has long been a very active research area. Traditional search-based algorithms (e.g. Dijkstra and tree-based searches)~\cite{kleinberg2006algorithm} convert a given physical space into an abstracted graph structure, where nodes represent locations and edges connections. The edges are further defined with weights that reflect certain travel costs. Then, given a set of nodes and some travel constraints, a search algorithm returns desired routes that can traverse through the queried nodes. While such exact search algorithms can provide very optimal solutions, they often do not scale up well. Therefore, heuristic search algorithms (e.g. A*\cite{lesterPathfindingBeginners}, LPA*~\cite{koenig2004lifelong}) and machine learning methods based on approximate optimization have emerged. The main advantage is that they typically avoid processing global or exact information, and thus end up with an acceptable sub-optimal solution but with a significant improvement in search efficiency. More recently, route recommendation algorithms have experienced a revolution brought by deep learning. With the advances in data storage and computing technologies, deep learning~\cite{lecun2015deep} prevails and has demonstrated much stronger data representation and knowledge extraction abilities. It enables dealing with more complex route recommendation problems that are large-scale, multi-objective, and personalized-oriented with multi-source, multi-modal contextual information~\cite{wang2020deep}. Therefore, how the classic methods and the modern methodologies are inherently connected and improved, and what novel applications have been conceived need a thorough review from new perspectives.

Our survey of route recommendation mainly focuses on the work that is conducted within the context of urban computing~\cite{zheng2014urban}, where most of the applications flourish. To start, we first searched for survey articles that are broadly related to route recommendation, as shown in \tableautorefname~\ref{tab:related review}. But we find that most surveys have only briefly mentioned urban computing scenarios for route recommendation, without providing a deeper understanding of the data and methods. The most related surveys to our scope of review are by Siriaraya~\textit{et al.}~\cite{siriarayaShortestRouteSurvey2020} and Li~\textit{et al.}~\cite{liRouteSearchPlanning2021}. Siriaraya~\textit{et al.}~\cite{siriarayaShortestRouteSurvey2020} introduced route recommendation from the perspective of navigation systems and classified them based on their proposed SWEEP classification standard. It also provided an overview of various data sources, algorithms, and evaluation methods for implementing quality-aware path navigation systems. Li~\textit{et al.} mainly focused on search-based methods for route recommendation tasks from the perspective of trajectory data~\cite{liRouteSearchPlanning2021}. It elaborated on data storage techniques for trajectories and presented a few matching methods among trajectories, thus having touched on the fundamentals of route recommendation. However, none of the abovementioned surveys concentrate on modern route recommendation work, so our survey aims to fill this gap. Most of the papers reviewed in this survey were published in well-known international conferences (e.g. SIGKDD, SIGSPATIAL, NeurIPS, ICLR, ICML, KDD, WWW, IJCAI, and VLDB) and journals (e.g. ACM or IEEE Transactions, Elsevier or Springer journals) in various research fields of artificial intelligence, big data, data mining, and network.

\begin{figure*}[hbt]
	\centering 
	\includegraphics[width=0.8\textwidth]{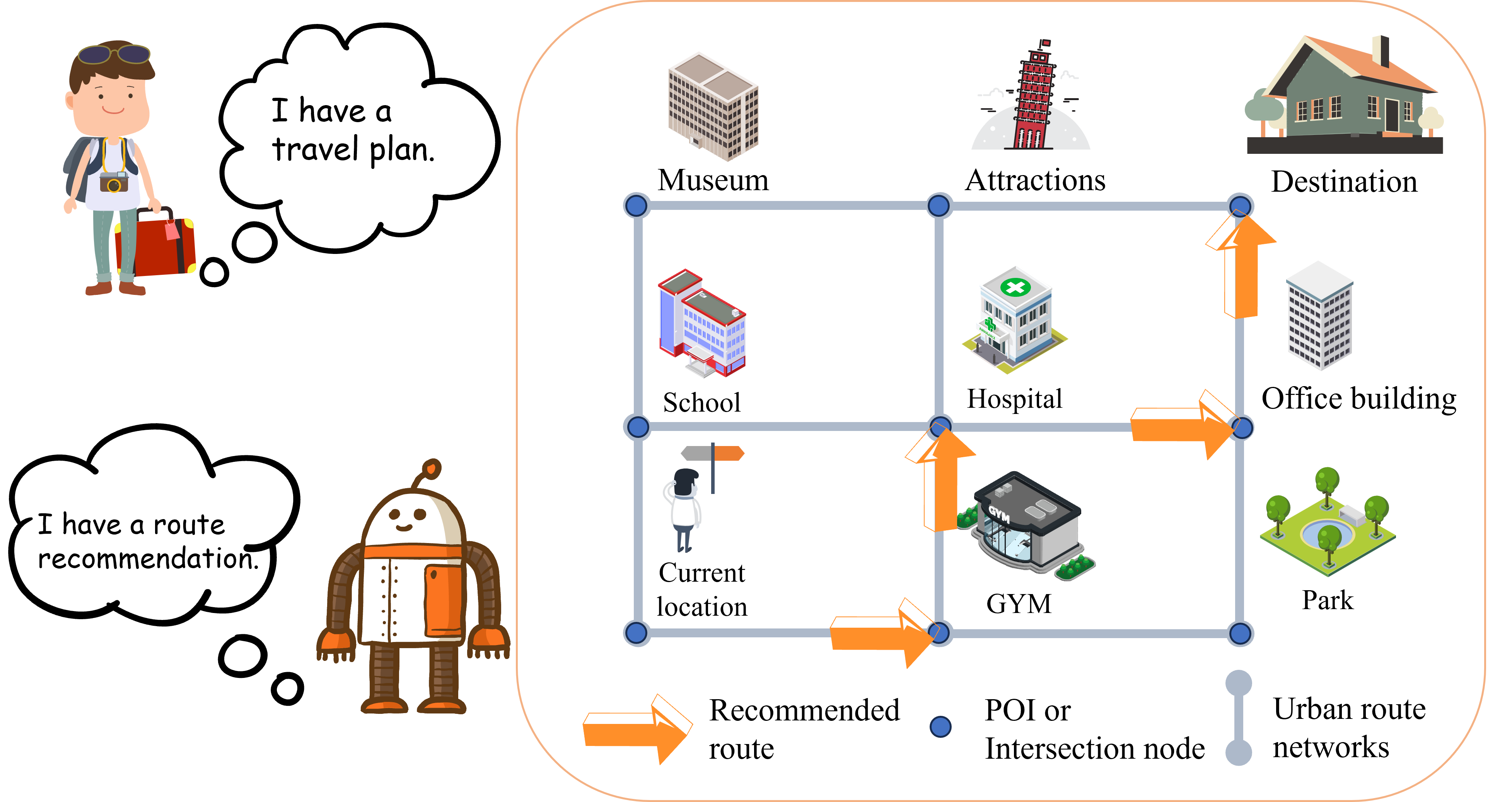}	
    \caption{A route recommendation task. A user specifies a travel plan defined by departure time, source location, destination, and other preferences, and an algorithm outputs a desired route recommendation that satisfies the query.}
	\label{fig:routerecommendation}%
\end{figure*}

Our survey provides a comprehensive review of route recommendation methods and applications based on urban computing scenarios~\cite{zheng2014urban}. We survey existing work from both classic methods and modern deep learning perspectives. Our contributions are threefold: 
\begin{itemize} 
\item[1.] We present an overview of route recommendation research work, from its early development to the current state and future trends. We stand at the viewpoint of urban computing~\cite{zheng2014urban}, which is a new interdisciplinary field that aims to improve urban life with big data and ubiquitous computing. 
\item[2.] We categorize and compare a wide range of methodologies and applications of route recommendation. We cover both classical machine learning methods such as collaborative filtering, matrix factorization, and Markov models as well as modern deep learning approaches such as graph neural networks, attention mechanisms, and generative models. We also demonstrate many recent novel applications of route recommendation, e.g., personalized travel planning, traffic optimization, and urban exploration. 
\item[3.] We summarize existing challenges and limitations of route recommendation and envision several promising future directions. For example, incorporating multi-modal data, enhancing user privacy, exploiting social and environmental factors, and integrating with large models. \end{itemize}


The structure of this survey is organized below.
Section~\ref{sec:Background} gives a brief introduction of the background, problem definitions, evaluation metrics, and datasets for route recommendation. Section~\ref{sec:Traditional} overviews classic methods route recommendation methods. Section~\ref{sec:Deep} reviews and classifies modern route recommendation works based on deep learning. Following that, Section~\ref{sec:Application} demonstrates different categories of route recommendation applications. And finally, Section\ref{sec:Conclusion} discusses current open problems and points out the future research directions.

\section{Background}
\label{sec:Background}
In this section, we provide an overview of the background. Route recommendation is a crucial function in various applications such as navigation systems, logistics, and travel planning. In our survey, we stand from the perspective of urban computing, which is a new interdisciplinary field that leverages big data and ubiquitous computing to improve urban life. In the following, we formally introduce the definitions, algorithms, and evaluation metrics of route recommendation.

\subsection{What is Route Recommendation?}
A route recommendation task aims to provide a suitable route for a user based on their travel constraints and preferences. \figurename~\ref{fig:routerecommendation} illustrates an example of this task. Here, a user needs to visit a hospital nearby and then continue to stop by a friend’s home. A route recommendation model takes such a query as input and outputs one or more desired routes that satisfy the user’s requirements. This is a basic route recommendation process, but more complicated queries can be further extended. Below are formal definitions associated with route recommendation.


\begin{figure}[h]
	\centering 
	\includegraphics[width=0.3\textwidth]{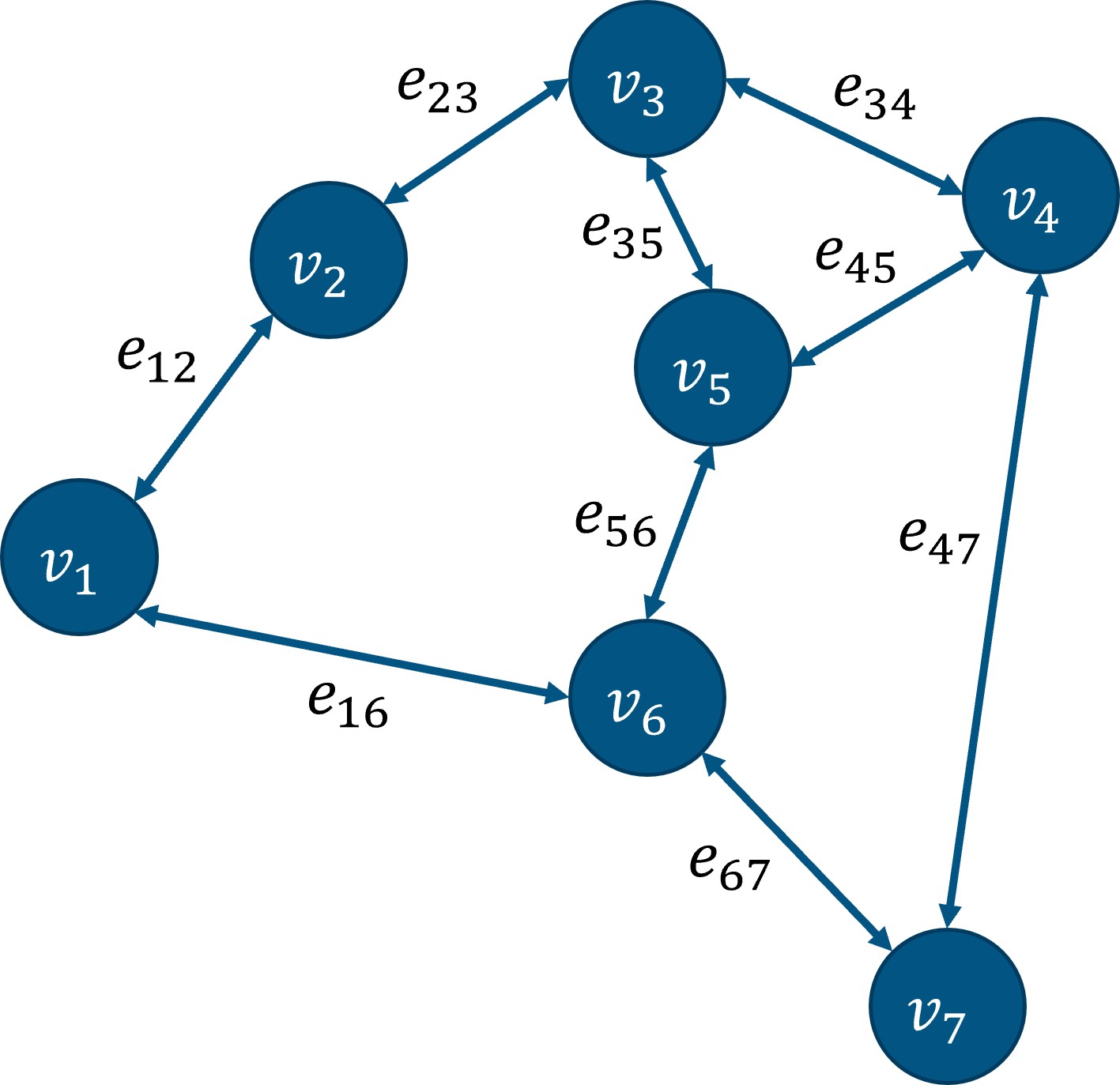}	
	\caption{A graph abstraction of a road network.} 
	\label{fig:road_network}%
\end{figure}

\begin{definition}
A road network is a directed graph $G = <V, E, I>$, where $V$ represents a set of nodes, $E$ a set of edges, and $I$ the graph information. Let $|V|$ represent the number of nodes, and $|E|$ the number of all edges. Then, $V=\{v_1, v_2, \dots, v_{|V|}\}$ and each $e_{ij} \in E$ represents an edge that directly connects $v_i$ and $v_j$. $e_{ij}$ can be a scalar or a vector storing different types of weights. $I$ optionally contains the graph information such as additional edge information, transitional probabilities, and node features.
\end{definition}

\figurename~\ref{fig:road_network} shows a simple urban traffic road network with seven nodes and nine edges. In this example, all edges are bidirectional and only one direction is marked out. 
Note that $G$ is built from knowledge and/or from data. Minimally, $G$ should have $V$ and $E$ defined; but more recently, graphs are built from various types of data, and thus a graph may contain much richer information. As a result, the basic definition of $G=<V, E>$ becomes insufficient, so we add a generic field $I$ that can store such additional graph information.

\begin{definition}
\label{def2}
    A route $r$ is defined as a sequence of directly connected nodes. $r(s, d)=[v_{(1)},v_{(2)},\dots,v_{(k)}]$, where $v_{(1)}=s$ is the starting node, and $v_{(k)}=d$ is the destination node. Note that if $v_{(i)}=v_x$ and $v_{(i+1)}=v_y$, then it implies that $e_{xy} \in E$ (direct connection). We further denote all the routes traversing from node $s$ to node $d$ as $R(s, d)$, i.e. $r(s, d) \in R(s, d)$.
\end{definition}

We define routes by graph nodes, as shown in Definition \ref{def2}. Other literature also uses edges to represent routes. For example, the route $r=\{v_1, v_6, v_7, v_4\}$ in \figurename~\ref{fig:road_network} can be written as $r=\{e_{16}, e_{67}, e_{74}\}$.

\begin{definition}
\label{def3}
    A user's query is defined as a tuple $q=<s, d, o>$, where $s$ and $d$ represent the start node and the end node respectively, and $o$ includes other optional information like travel preferences. We use $q.s$, $q.d$, and $q.o$ to denote the corresponding fields in $q$.
\end{definition}

A route recommendation algorithm (Definition \ref{def4}) aims to find the optimal route(s) that meets the user’s query $q$ (Definition \ref{def3}). Besides $q.s$ and $q.d$, the optional information $q.o$ can contain departure time, a set of waypoints, a maximum travel cost, and so on. Such additional information often serves as the constraints of the optimization objectives.

\begin{definition}
\label{def4}
    A route recommendation algorithm (or model) returns the most probable route $r^*$ that satisfies a user's query $q$ based on a network $G$:
    $$r^*=\mathop{\arg\max}\limits_{r \in R(q.s,\ q.d)}P(r \mid q, G)$$
\end{definition}

Definition \ref{def4} defines the optimality from the probabilistic view, though other types of optimality may also be adopted (see subsection \ref{subsec:eva_metr}). The probabilistic measure of optimality is widely used and is a natural choice for many modern methodologies, most of which are \textit{data-driven} methods. Hence, the probabilistic formulation can be defined based on data, for example, traffic flows among a network.
Given such a setting, often by assuming appropriate probabilistic independence, the probability of the whole sequence of $r$ can be decomposed as multiplications of step-wise probabilities. As shown by Equation (\ref{P1}), $P(r)$ can be computed as the multiplication of transitional probabilities $P(v_{(i+1)} \mid v_{(i)})$, which are pre-defined in the given graph $G$.

\begin{equation}
    P(r \mid q, G)=\prod_{i=1}^{k-1} P\left(v_{(i+1)} \mid v_{(i)}; q, G \right)
\label{P1}
\end{equation}
Similarly, when using edges to represent routes, Equation (\ref{P1}) can be equivalently rewritten as follows:

\begin{equation}
    P(r \mid q, G)=\prod_{i=1}^{k'-1} P\left(e_{(i+1)} \mid e_{(i)}; q, G \right)
\label{P2}
\end{equation}

To find $r^*$, one can compute all the $P(r)$ for every $r \in R(s, d)$ and return the optimal one. However, this can yield factorial computation complexity. In practice, people often adopt dynamic programming to solve this problem (even the global optimality may no longer hold). That is, finding the optimal $r^*$ is reduced to finding a sequence of step-wise optimals. Formally, let $v^*_{(1)}=s$, then the next optimal node $v^*_{(i+1)}$ transitioning from $v^*_{(i)}$ is the one that has the maximum transitional probability:

\begin{equation}
    v^*_{(i+1)}=\mathop{\arg\max}\limits_{v \in \{v^*_{(i)}\text{'s neighbor}\}}P(v \mid v^*_{(i)}; q, G)
\label{rsdmaxv}
\end{equation}
Then, the optimal nodes $[v^*_{(1)}, v^*_{(2)}, \dots]$ comprise the optimal $r^*$:
\begin{equation}
    r^*=\mathop{\arg\max}\limits_{[v_1, v_2, \dots]}\prod_{i}P(v_{(i+1)} \mid v^*_{(i)}; q, G)
\label{rsdmaxr}
\end{equation}
To optimize the above maximum probability function, one can transform it into a sum of negative log probabilities and find the minimal:
\begin{equation}
    r^*=\mathop{\arg\min}\limits_{[v_1, v_2, \dots]}\sum_{i}-\log P(v_{(i+1)} \mid v^*_{(i)}; q, G)
\label{rsdmaxe}
\end{equation}



\subsection{Evaluation Metrics}
\label{subsec:eva_metr}
To evaluate routes returned by an algorithm, there are a few commonly used metrics. Most of them are \textit{supervised}, meaning that an algorithm-found route $r^*$ is compared to $\tilde{r}$, which is a \textquoteleft true label\textquoteright\ or \textquoteleft golden route\textquoteright, defined differently by situations.
Note that these metrics can also be used to define optimality in a model, provided the training data are labeled.

\noindent\textbf{Edit distance}. Edit distance, or \textit{Lev} distance, named after Levenshtein, was proposed by Ristad and Yianilos~\cite{ristad1998learning}, which refers to the minimum number of editing operations required to convert one string to another. In our setting, it can be used to measure the difference between $r^*$ and $\tilde{r}$. In Definition \ref{def2}, a route is defined as a sequence of nodes, so we denote by $r[i]$ ($1 \leq i \leq k$) the $i$-th node of $r$. Let the edit distance between $r^*$ and $\tilde{r}$ be denoted by $\operatorname{Lev}(r^*, \tilde{r})$. One can use dynamic programming to compute $\operatorname{Lev}(r^*, \tilde{r})$, of which the recursive calculation is defined as follows:

    \begin{footnotesize}
    \begin{equation}
    \operatorname{Lev}_{r^*, \tilde{r}}(i, j)=
    \begin{cases}
    \max (i, j) & \text{if } i*j=0 \\
    \min \begin{cases}\operatorname{Lev}_{r^*, \tilde{r}}(i-1, j)+1 \\ \operatorname{Lev}_{r^*, \tilde{r}}(i, j-1)+1 \\ \operatorname{Lev}_{r^*, \tilde{r}}(i-1, j-1)+1_{\left(r^*[i] \neq \tilde{r}[j]\right)}\end{cases} & \text {otherwise}
    \end{cases}
    \label{EDT}
    \end{equation}
    \end{footnotesize}

\noindent\textbf{Precision}. Precision measures the percentage of the number of overlapped nodes between the nodes in $r^*$ and the nodes in $\tilde{r}$ over the number of nodes in $\tilde{r}$. Let $r^* \cap \tilde{r} = \{v \mid v \in r^* \wedge v \in \tilde{r}\}$, then precision is defined as:
    \begin{equation}
    \text { Precision }=\frac{\left|r^* \cap \tilde{r}\right|}{\left|\tilde{r}\right|}
    \label{eq:precision}
    \end{equation}

\noindent\textbf{Recall}. Recall measures the percentage of the number of overlapped nodes over the number of nodes in $r^*$ instead:
    \begin{equation}
    \text { Recall }=\frac{\left|r^* \cap \tilde{r}\right|}{\left|r^*\right|}
    \label{eq:recall}
    \end{equation}

\noindent\textbf{F1 score}. F1 score simultaneously considers precision ($P$) and recall ($R$) and is in fact a harmonic average of precision and recall:
    \begin{equation}
    F_1=\frac{2 * P * R}{P+R}
    \label{eq:f1score}
    \end{equation}

Apart from the above metrics, in real-world application scenarios such as taxi revenue, travel route recommendations, etc., there will be more particular ways to evaluate results, and those will be discussed inline with certain applications in Section ~\ref{sec:Application}.


\begin{table*}[ht]
\renewcommand{\arraystretch}{1.5}
\caption{Route recommendation datasets.}
	\centering
    \small
	\begin{tabularx}{\textwidth}{p{2.5cm}p{5cm}X}
        \toprule[2pt]
        \textbf{Domain}&\textbf{Dateset \& Link}&\textbf{About}\\
        \midrule
        \multirow{3}{*}{Route Networks}& \href{https://www.openstreetmap.org/}{City and Country Maps} & \multirow{3}{*}{Intersection IDs, road structure, POI, subway stations, etc.} \\  
                    & \href{https://sites.google.com/ualberta.ca/nascimentodatasets/}{Mario A. Nascimento}~\cite{AhmadiDatasets} &  \\ 
                    & \href{https://publish.illinois.edu/dbwork/open-data/}{New York Road Network}~\cite{chenEffectiveEfficientReuse2019} & \\ \cline{2-3}

        \multirow{4}{*}{Security-related} & \href{https://data.cityofchicago.org/stories/s/Crimes-2001-to-present-Dashboard/5cd6-ry5g}{Chicago Crime Record}~\cite{islamPrivacyEnhancedPersonalizedSafe2021} &  Criminal records, including  regions, types of crimes,    \\
                        & \href{https://opendataphilly.org/datasets/crime-incidents/}{Philadelphia Crime Record}~\cite{islamPrivacyEnhancedPersonalizedSafe2021}& pursuit situations, etc. Meteorological Agency data mainly \\
                        & \href{https://www.ecns.cn/cns-wire/2013/07-12/72886.shtml}{Beijing Crime Hot Spots}~\cite{islamPrivacyEnhancedPersonalizedSafe2021} & include information such as weather, disaster types, \\ 
                        & \href{https://www.jma.go.jp/jma/menu/menureport.html}{Japan Meteorological Agency}~\cite{biEvacuationRouteRecommendation2019}& longitude, and latitude. \\ \cline{2-3}
                        
         \multirow{4}{*}{Tourism-related} &\href{https://github.com/raingo/yfcc100m-entity}{YFCC100M}~\cite{gaoAdversarialHumanTrajectory2023} &  \\
        &\href{https://data.cityofnewyork.us/City-Government/Points-Of-Interest/rxuy-2muj}{New York Road POI}~\cite{chenEffectiveEfficientReuse2019} &POI Pictures and corresponding text comments, geographical \\
        &\href{http://www.nyc.gov/html/tlc/html/home/home.shtml}{NYC TCL record}~\cite{sunOnlineDeliveryRoute2019}&location information of scenic spots. \\
        & \href{https://www.matsim.org/}{MATSim}~\cite{TransportSimulationMATSim2016}&\\ \cline{2-3}
        
         \multirow{2}{*}{Indoor-related}    &  \href{https://plans.trafimage.ch/}{Station plans from Swiss federal railway SBB}~\cite{liBarrierFreePedestrianRouting2021}  & Indoor-related datasets, including information such as store IDs, floors, and locations. They contain information such \\
        &\href{ https://longaspire.github.io/s/hkdata.html}{Indoor Venue Keyword Dataset}~\cite{feng2020indoor}&as brand names, store sizes, types, etc. \\  \cline{2-3}
        Constraint-related &\href{http://www.diag.uniroma1.it/challenge9/download.shtml}{New York transportation network}~\cite{wangConstrainedRoutePlanning2021}  & New York City bus and subway stations, public transportation routes, and schedules. \\ \cline{2-3}
         \multirow{5}{*}{Economics-related} & \href{https://tianchi.aliyun.com/competition/entrance/231777/information }{Food Pick-up and Delivery Data}~\cite{wenGraph2RouteDynamicSpatialTemporal2022}&  \\ 
        & \href{https://github.com/lucktroy/DeepST/tree/master/data/TaxiBJ}{Taxi-BJ} &This mainly includes information such as taxi ID, driving    \\ 
        & \href{https://learn.microsoft.com/zh-cn/azure/open-datasets/dataset-catalog}{Taxi-NYC}~\cite{zheng2015trajectory} & trajectory, passenger status, and speed. \\
        &\href{https://gaia.didichuxing.com/}{DiDi}& \\
        &\href{https://tianchi.aliyun.com/dataset/94216}{Porto Taxi}~\cite{jainNeuroMLRRobustReliable2021}&  \\

		\bottomrule[2pt]
	\end{tabularx}%
	\label{tab:Datasets}%
	
\end{table*}%

\subsection{Datasets}

Lastly, we present common datasets relevant to route recommendation. Since route recommendation tasks mainly use trajectory data and road network information, additional auxiliary data such as passenger status and scenic spots are often included as well. As some datasets are essentially private and require access permissions, we here mainly list publicly accessible datasets, shown in ~\tableautorefname~\ref{tab:Datasets}. The list has included map data, trajectory data, POI and attraction data, vehicle speed data, etc. The categories are chosen consistent to the application scenarios to be presented in Section \ref{sec:Application}.

In the list, road network data include the basic construction of urban maps and the calibration of some POIs. For example, the Mario dataset covers the intersection IDs and latitudes within a city, with the edge data representing the road structure and connectivity of the city. POI data include information such as the location of well-known buildings, bus stops, and subway stations in a city.
Security-related data contain criminal records and natural disaster records. Among them, criminal records include different regions, types of crimes, pursuit situations, etc. Natural disaster data mainly include information such as weather, disaster types, longitude, and latitude.
Tourism-related data mainly include pictures and corresponding text comments of scenic spots, as well as geographical location information of scenic spots.
Indoor-related data cover information such as store IDs, floors, and locations. They can further contain brand names, store sizes, types, etc.
Constraint-related data offer information on New York City bus and subway stations, public transportation routes, and schedules.
Economics-related data mainly focus on taxis. This includes information such as taxi ID, driving trajectory, passenger status, and speed. It can reflect the income level of taxis over a while and better fit the revenue objective function to calculate route recommendation results.

\begin{table*}[h]
\renewcommand{\arraystretch}{1.5}
\caption{Classification of Route Recommendation methods}
	\centering
    \small
	\begin{tabularx}{\textwidth}{Xp{5cm}p{6cm}}
		\toprule[2pt]
	  \textbf{Category of methods}& \textbf{Refined classification}& \textbf{References}  \\
		\midrule
		\multirow{7}{*}{Classic Methods}  &     Search-based methods      & \cite{drozdek2012data}~\cite{EBENDT20091310}~\cite {likhachev2003ara}~\cite{stentz1994optimal}~\cite {koenig2002d}~\cite {sanders2005highway}~\cite {sanders2006engineering}~\cite{geisberger2008contraction}~\cite{delling2011customizable}
  \cite{xuTravelRouteRecommendation2016}~\cite{mohdnordinApplicationAlgorithmAmbulance2011}~\cite{gareauEfficientElectricVehicle2019}~\cite{ahmadiBestCompromiseInRouteNearest2017}~\cite{chenEffectiveEfficientReuse2019}~\cite{chondrogiannisFindingKDissimilarPaths2018}~\cite{hackerMostDiverseNearShortest2021}~\cite{costaOnlineInRouteTask2020}~\cite{chenOnlineRoutePlanning2021}
  \cite{fitzgeraldOnlineRouteReplanning2021}~\cite{daiPersonalizedRouteRecommendation2015}~\cite{wangR3RealtimeRoute2014}~\cite{costaInrouteTaskSelection2018}~\cite{lesterPathfindingBeginners}~\cite{koenig2004lifelong}\\ \cline{3-3}
            &         Probability-based methods        & \cite{lavalle1998rapidly}~\cite{lavalle2001randomized}~\cite{kuffner2000rrt}~\cite{karaman2011sampling}~\cite{chenLearningPointsRoutes2016} ~\cite{quProfitableTaxiTravel2020}~\cite{qianSmartRecommendationMining2012}~\cite{rajanTieringContractionEdge2021}\\ \cline{3-3}
            &        Biomimetic methods       &\cite{zhengNovelMultiObjectiveMultiConstraint2021}~\cite{liuRealtimePersonalizedRoute2014}~\cite{nguyenACObasedTrafficRouting2023} ~\cite{liangRouteRecommendationBased2022}\\ \cline{3-3}
            &        Clustering-based methods       &\cite{chenPersonalizedNavigationRoute2023}~\cite{liPhysicsguidedEnergyefficientPath2018}~\cite{BPlanner2014}~\cite{SmartZhou2019}~\cite{louAnalysis2022}~\cite{Yuco2023}~\cite{Chenper2021} \\ \cline{3-3}
             &         Constraint-based methods &\cite{quCosteffectiveRecommenderSystem2014}~\cite{salgadoEfficientApproximationAlgorithm2018}~\cite{huangBackwardPathGrowth2015}~\cite{liBarrierFreePedestrianRouting2021}~\cite{barthCYCLOPSCYCLeRoute2018}~\cite{liPhysicsguidedEnergyefficientPath2018}~\cite{tengSemanticallyDiversePaths2021}~\cite{chengTaxiCTaxiRoute2019}~\cite{rezaOptimalRouteStops2017}
             \cite{rajanPhaseAbstractionEstimating2019}~\cite{laiUrbanTrafficCoulomb2019}~\cite{herschelmanConflictFreeEvacuationRoute2019}\\ \cline{2-3}
         \multirow{7}{*}{Deep Learning Methods}   &       Hybrid methods &     \cite{liu2021ldferr}~\cite{wangASNNFRRTrafficawareNeural2023}~\cite{wangEmpoweringSearchAlgorithms2019}~\cite{wuLearningEffectiveRoad2020}~\cite{huangMultiTaskTravelRoute2021}~\cite{jainNeuroMLRRobustReliable2021}~\cite{yangNoahNeuraloptimizedSearch2021}~\cite{agostinelliObtainingApproximatelyAdmissible}~\cite{wangPersonalizedRouteRecommendation2022}~\cite{liuIntegratingDijkstraAlgorithm2020}  \\ \cline{3-3}
             &      Sequence-based methods      &\cite {wangASNNFRRTrafficawareNeural2023}~\cite{wangEmpoweringSearchAlgorithms2019}~\cite{wenGraph2RouteDynamicSpatialTemporal2022}~\cite{huangMultiTaskTravelRoute2021}~\cite{wangPersonalizedRouteRecommendation2022}~\cite{hahnPredictiveCollisionManagement2020}~\cite{liu2021ldferr}~\cite{fuProgRPGANProgressiveGAN2021}~\cite{zhangWalkingDifferentPath2018}
             \cite{biSpeakNavVoicebasedNavigation2021} \cite{BhumikaMARRS} \cite{WangPersonalizedLong2022} \cite{WangPersonalizedSpecified2021} \cite{WangQuery2Trip2023} \cite{HoPOIBERT2022}\\ \cline{3-3}
                 &    Graph-based methods      & \cite{wangASNNFRRTrafficawareNeural2023}~\cite{gaoDualgrainedHumanMobility2023}~\cite{wangEmpoweringSearchAlgorithms2019}~\cite{wenGraph2RouteDynamicSpatialTemporal2022}~\cite{wuLearningImprovementHeuristics2022}~\cite{wuLearningEffectivelyEstimate2019}~\cite{wuLearningEffectiveRoad2020}~\cite{jainNeuroMLRRobustReliable2021}~\cite{yangNoahNeuraloptimizedSearch2021}
                 \cite{wangPersonalizedRouteRecommendation2022} \\ \cline{3-3}
                  &    Multi-modal methods     & \cite{zhangWalkingDifferentPath2018}~\cite{biSpeakNavVoicebasedNavigation2021}~\cite{parkStudyTopicModels2022}~\cite{padia2019sentiment}~\cite{huAGREEAttentionBasedTour2019}~\cite{yuAdvancedMultimodalRouting2012}~\cite{zhangPersonalizedMultimodalRoute2014}~\cite{bucherHeuristicMultimodalRoute2017}~\cite{campigottoPersonalizedSituationAwareMultimodal2017}
                  \cite{herzogRouteMeMobileRecommender2017}~\cite{chengTERPTimeEventDependentRoute2019}~\cite{liuJointRepresentationLearning2019}~\cite{georgakisMultiModalRoutePlanning2019}~\cite{liuHydraPersonalizedContextAware2019}~\cite{liu2020multi}~\cite{liuIncorporatingMultiSourceUrban2022}
                  \cite{liuUnifiedRouteRepresentation2023b} \\ \cline{3-3}
                 &    Reinforced learning methods      &\cite{xiaEfficientNavigationConstrained2022}~\cite{chenCuRLGenericFramework2023}~\cite{biEvacuationRouteRecommendation2019}~\cite{liuIntegratingDijkstraAlgorithm2020}~\cite{wuLearningImprovementHeuristics2022}~\cite{agostinelliObtainingApproximatelyAdmissible}~\cite{liuPersonalizedRouteRecommendation2022}~\cite{jispatiotemporalFeatureFusion2020}\\ 

		\bottomrule[2pt]
	\end{tabularx}%
	\label{tab:Method}%
	
\end{table*}%

\section{Classic Methods for Route Recommendation}
\label{sec:Traditional}
We start with reviewing classic route recommendation algorithms. This section mainly covers non-deep-learning methods which usually have a long history, so we name them as \textit{Classic Methods} (CM). 
Although classic, they still lay the foundation for many modern methods.
Typically, CM aims to find optimal routes according to a given objective (often single-objective), such as minimum distance, time, or cost. These algorithms, methods, or models can be roughly categorized into five groups: search-based, probability-based, biomimetic-based, clustering-based, and constraint-based. Many of these methods originated from the field of robotics, where automatic pathfinding tasks are crucial. Later, they were adapted and applied to various other domains such as urban road finding and navigation. In the following subsections, we will discuss the representative methods and their extensions for each group.

\subsection{Search-based Methods}
\label{subsec:search}
\begin{figure}[h]
		\centering
		\includegraphics[width=0.8\linewidth]{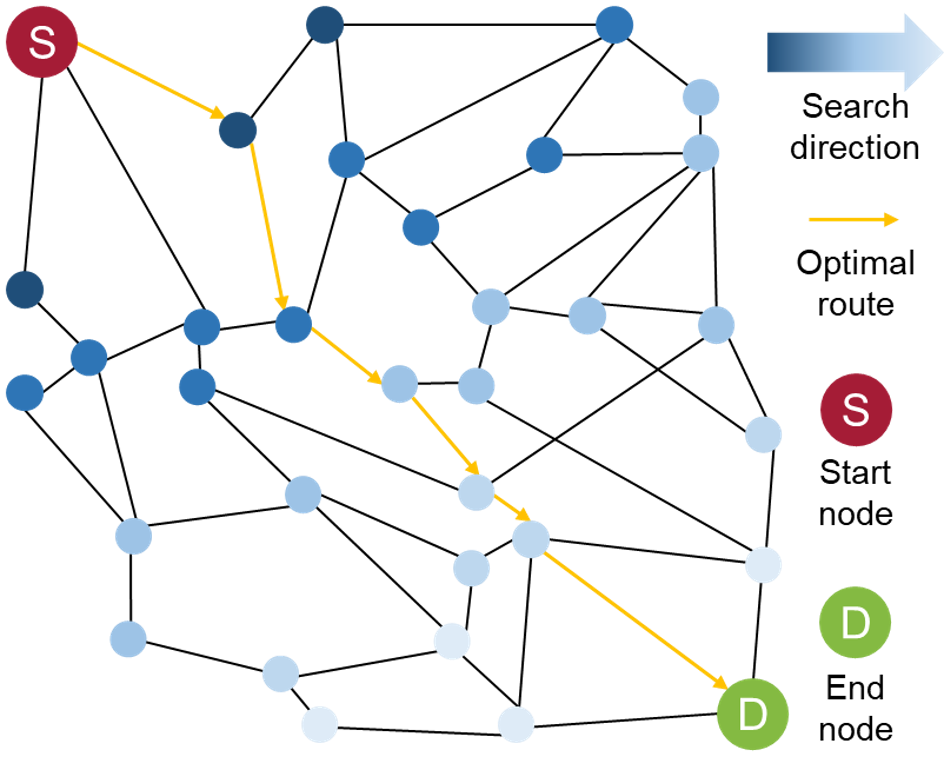}
		\caption{An example process of search-based methods. Featured by strategically searching for available paths that connect the starting node and the end node, and finally returning the computed optimal routes.}
		\label{fig:searchbased}
\end{figure}

Search-based methods work by exploring all possible paths from one node to the other, and meanwhile, filtering out the paths that fail to satisfy the given constraints such as a maximum distance or cost. A search tree can often be built to construct the search space, and many search algorithms may apply, such as recursive search, dynamic programming, branch and bound method, etc. Search-based algorithms are usually simple and optimal but can also be inefficient. They are not suited to deal with large networks with dynamic changes. An example process of the search-based method is shown in Figure~\ref{fig:searchbased}.

The early concept of route recommendation originated from data structure textbooks. For instance, Drozdek~\cite{drozdek2012data} introduced breadth-first search (BFS) and depth-first search (DFS) algorithms, which are the basic search methods for graph-structured data. Moreover, to find the optimal paths that achieve a predefined criterion (e.g. shortest path), Floyd’s algorithm and Dijkstra’s algorithm are the classic methods, but they also suffer from a computational complexity of $O(N^3)$ for finding optimal paths between any pair of nodes in a graph.

The abovementioned algorithms are called \textit{exact} methods since they traverse through all the nodes and paths. However, this can be computationally intractable for large networks. To avoid exhaustive searches, Lester~\cite{lesterPathfindingBeginners} proposed A*, which employs a heuristic search function that guides searches to the optimal directions. For example, consider a cost function $F(x)=g(x)+h(x)$ for finding a shortest path, where $g(x)$ is the shortest distance traveled from the source node $v_s$ to $v_x$, and $h(x)$ is the add-on heuristic function that estimates the distance from $v_x$ to the destination node $v_d$. Having such an $h(x)$, the algorithm can avoid finding the smallest $g(x)$ for every node but is instead, guided to find those nodes with small $h(x)$. Despite this efficiency improvement, one must be warned that good heuristics may not be easy to design as it requires good \textquoteleft pre-knowledge\textquoteright\ of the problem.

So far, we have implicitly assumed static graphs. But in practice, urban road accessibility can change over time due to road work, accidents, public events, etc. How to dynamically generate an alternative path is critical to real-time applications. To address this problem, Ebendt and Drechsler~\cite{EBENDT20091310} proposed a weighted A* algorithm, which considers the difference between the estimated distance and the actual distance, and they trade off the optimality of A* for faster computation speed and real-time route recommendation. Likhachev~\textit{et al.}~\cite {likhachev2003ara} developed ARA*, which is another weighted A* algorithm. They iteratively compute the optimal path by search step and optimization step until the weighted value $\alpha$ is reduced to one.
Stentz~\cite{stentz1994optimal} proposed the D* algorithm that improves A* in real-time situations. The D* algorithm maintains the latest state information for nodes and path accessibility for edges such that the recommended paths are both accessible and optimal. 
Koenig~\textit{et al.}~\cite{koenig2004lifelong} introduced the LPA* algorithm based on A*. The LPA* algorithm can adapt to changes in the graph by updating the $g$ value during every search period and making necessary corrections. Similar to A*, LPA* also uses heuristic functions that are derived from the lower bound cost of the path from a given node to the destination. 
Later, Koenig and Likhachev~\cite {koenig2002d} improved the LPA* algorithm by using dynamic programming to adapt to sudden changes of urban road conditions.

The rapid expansion of cities and urban road networks imposes an increasing demand for real-time computing. Therefore, efficient storage and computation methods for live searches are needed. Sanders and Schultes~\cite {sanders2005highway,sanders2006engineering} proposed the Highway Hierarchies (HH) algorithm optimized for practical applications. The main contribution is that HH reduces the size of network data when preprocessing directed graph structures, thus saving storage and improving computation speed. 
Geisberger~\textit{et al.}~\cite{geisberger2008contraction} introduced the Contraction Hierarchies (CH) algorithm, which searches and recommends the fastest path based on real-time travel time using heuristic functions. 
The system developed by Delling~\textit{et al.}~\cite{delling2011customizable} can support multiple types of optimization objectives for route recommendations in larger map structures, and their methods using overlay topology and parallel computing methods can speed up real-time searches and responses.

To close this subsection, we mention a few recent works that are characterized by the classic search-based methods.
Mohdnord~\textit{et al.}~\cite{mohdnordinApplicationAlgorithmAmbulance2011} applied the A* algorithm to the development of an ambulance routing system, which achieved faster and more efficient recommendations. 
Gareau~\textit{et al.}~\cite{gareauEfficientElectricVehicle2019} used path generation techniques to reduce the travel time of electric vehicles by searching for charging stations with less waiting time. 
Chen~\textit{et al.}~\cite{chenEffectiveEfficientReuse2019} developed two parallel algorithms, fully split parallel search (FSPS) and group split parallel search (GSPS), which used pruning techniques to deal with network expansions. 
Hacker~\textit{et al.}~\cite{hackerMostDiverseNearShortest2021} proposed an algorithm that searched over all the paths that satisfy the length constraints and generated their $k$ subsets as candidate results. 
Chen~\textit{et al.}~\cite{chenOnlineRoutePlanning2021} studied online route planning problems based on time-dependent road networks. They proposed two effective heuristic algorithms and accelerated them by using indexing techniques. 
Finally, Fitzgerald and Banaei-Kashani~\cite{fitzgeraldOnlineRouteReplanning2021} dealt with planning problems in a fully algorithmic routing scenario and proposed an algorithm that incorporated the impact of in-route replanning actions on the overall network performance.

\subsection{Probability-based Methods}
\label{subsec:prob}

\begin{figure}[h]
		\centering
		\includegraphics[width=\linewidth]{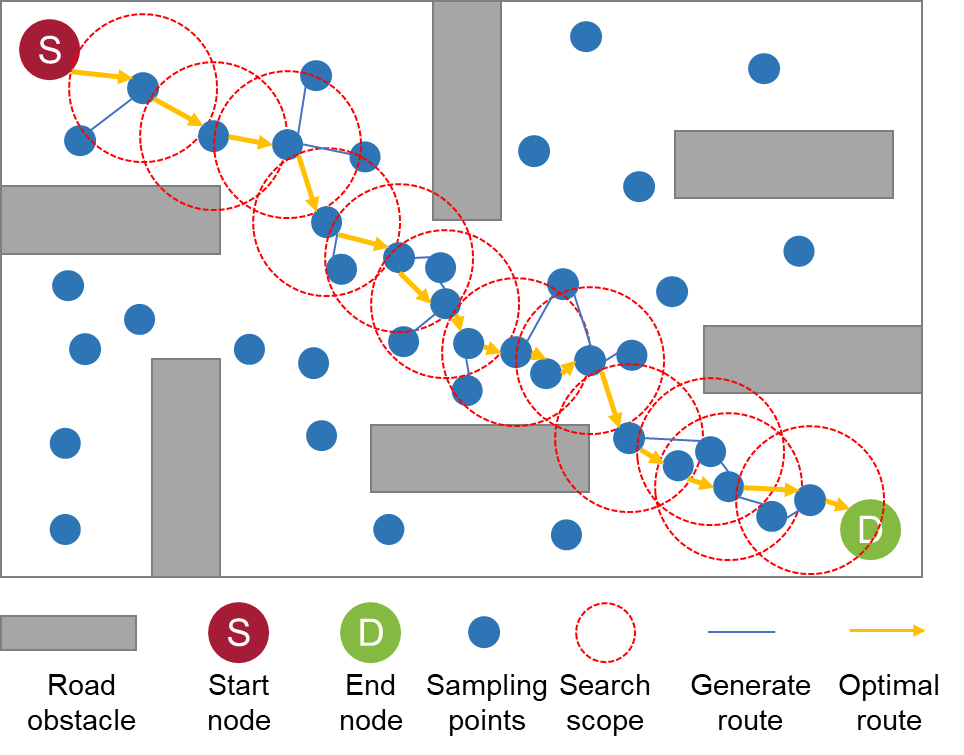}
		\caption{An example process of probability-based methods. Featured by obtaining the final route through sampling points and performing a proximal search in the map.}
		\label{fig:probabilitybased}
\end{figure}

Generally speaking, large graphs with complete sets of nodes and edges can still be hard to search, even enhanced with heuristics or pruning techniques. Therefore, sampling-based methods have been proposed, which essentially trade off search completeness for probabilistic completeness. In other words, graphs or routes can be sampled in part for construction. One example process is shown in Figure~\ref{fig:probabilitybased}. By sampling the passable areas on the map and searching for nearest neighbors, multiple passable paths are generated, and the optimal route is ultimately obtained through a search algorithm.

For example, Lavalle~\cite{lavalle1998rapidly} developed a Probabilistic Road Map (PRM), which constructs a path network in a complex environment by random sampling (transforms the continuous space into a discrete one) and then performs path planning on the sampled network to improve search efficiency in high-dimensional spaces. 
Besides, the following works, falling in the RRT family, are methods that sample paths.
First, Lavalle and Kuffner~\cite{lavalle2001randomized} proposed the Rapidly-exploring Random Tree (RRT) algorithm, which is a single-query algorithm that quickly finds a feasible path from the start point to the end point by random exploring. Its search process resembles a tree growing and expanding in different directions. 
Kuffner and LaValle~\cite{kuffner2000rrt} later introduced the RRT-Connect algorithm, which can be considered as a bidirectional expansion of two RRT trees. The two trees grow from the start/end point respectively, and they together generate a path when they connect.
Karaman and Frazzoli~\cite{karaman2011sampling} proposed RRT*, which is an asymptotically optimal algorithm. RRT* works similarly to RRT, except that RRT* has a reconnection process when selecting a parent node for a newly added node $v_{new}$. Instead of using the original parent node that generates $v_{new}$, RRT* selects a new parent $v_{pa}$ within a neighbor of $v_{new}$ that has the smallest path cost from $v_{pa}$ to $v_{new}$.

Several other works similarly used probabilistic models to recommend routes (particularly, driving paths in urban situations). 
Chen~\textit{et al.}~\cite{chenLearningPointsRoutes2016} adopted the Viterbi algorithm to find the most probable path with the highest random-walk probability and also used integer linear programming to avoid repeated traversal of the same paths. 
Qu~\textit{et al.}~\cite{quProfitableTaxiTravel2020} established a probabilistic network model using Kalman Filtering to predict the number of passengers in need of taxis at each location. They considered load balancing between passengers and taxis and introduced the shortest expected cruising distance to determine the potential cruising distance of taxis.
Qian~\textit{et al.}~\cite{qianSmartRecommendationMining2012} modeled the optimal driving process as a Markov Decision Process (MDP) and solved the MDP problem to obtain the optimal driving strategy for taxi drivers. 
Rajan and Ravishankar~\cite{rajanTieringContractionEdge2021} introduced a tiering technique for Contraction and Edge Hierarchies (CHs~\cite{geisberger2008contraction} and EHs~\cite{hespe2019more}) to deal with stochastic route planning problems. They developed Uncertain Contraction Hierarchies (UCHs) and Uncertain Edge Hierarchies (UEHs), which were shown to improve upon both CH and EH when processing real-time stochastic routing queries.

\subsection{Biomimetic-based Methods}
\begin{figure}[h]
		\centering
		\includegraphics[width=0.9\linewidth]{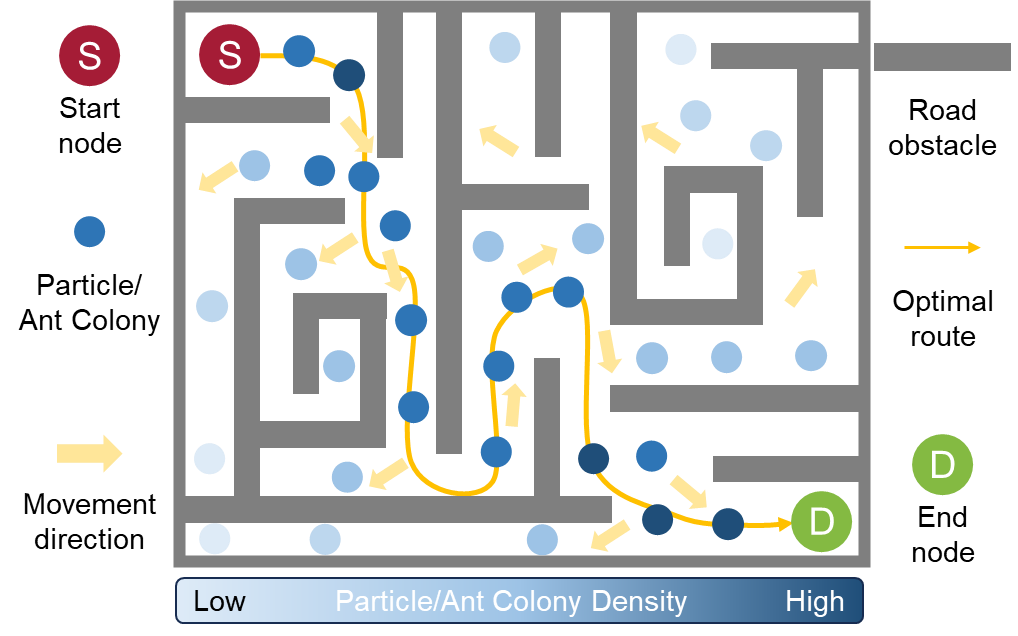}
		\caption{An example process of biomimetic-based methods. Featured by obtaining the final route based on the direction and speed of particle or ant colony movement. The particle density near the optimal route is marked high.}
		\label{fig:biomimeticbased}
\end{figure}

Biological organisms have inspired many artificial intelligence applications including path planning. Among them, one of the simplest forms of life is called physarum polycephalum, a slime mold that belongs to the amoeba species in the protozoa. This single-celled organism exhibits an interesting phenomenon: on a flat culture medium, its cellular growth path can provide valuable insights for the design of planar transportation networks~\cite{tero2010rules}. Therefore, methods that observe, summarize, and apply the rules of biological behaviors in reality to route recommendation are collectively called biomimetic methods. An example process is shown in Figure~\ref{fig:biomimeticbased}.

Several studies have applied swarm intelligence and genetic algorithms to personalized route recommendation. 
Zheng~\textit{et al.}~\cite{zhengNovelMultiObjectiveMultiConstraint2021} modeled the actual road network using Arc Map and constructed a personalized multi-constraint interest model based on an interest-label matching method and a scoring function. Then they proposed a neighborhood search algorithm and a hybrid particle swarm genetic optimization algorithm to recommend Top-K routes. 
Liu~\textit{et al.}~\cite{liuRealtimePersonalizedRoute2014} used genetic algorithms to simulate the users’ personalized path selection and the communication system between vehicles. 
Nguyen and Jung~\cite{nguyenACObasedTrafficRouting2023} proposed a decentralized traffic path system based on a new ant colony optimization algorithm pheromone model and designed an automatic negotiation technique in connected vehicle environments. They introduced a reverse pheromone model to generate repulsive forces between vehicles and provide negative feedback on congested roads. 
Liang~\textit{et al.}~\cite{liangRouteRecommendationBased2022} designed an efficient spatio-temporal metric-based routing recommendation scheme named RTS, which uses a simulated annealing algorithm to achieve real-time routing recommendations without much sacrifice of recommendation performance.


\subsection{Clustering-based Methods}

\begin{figure}[h]
		\centering
		\includegraphics[width=0.9\linewidth]{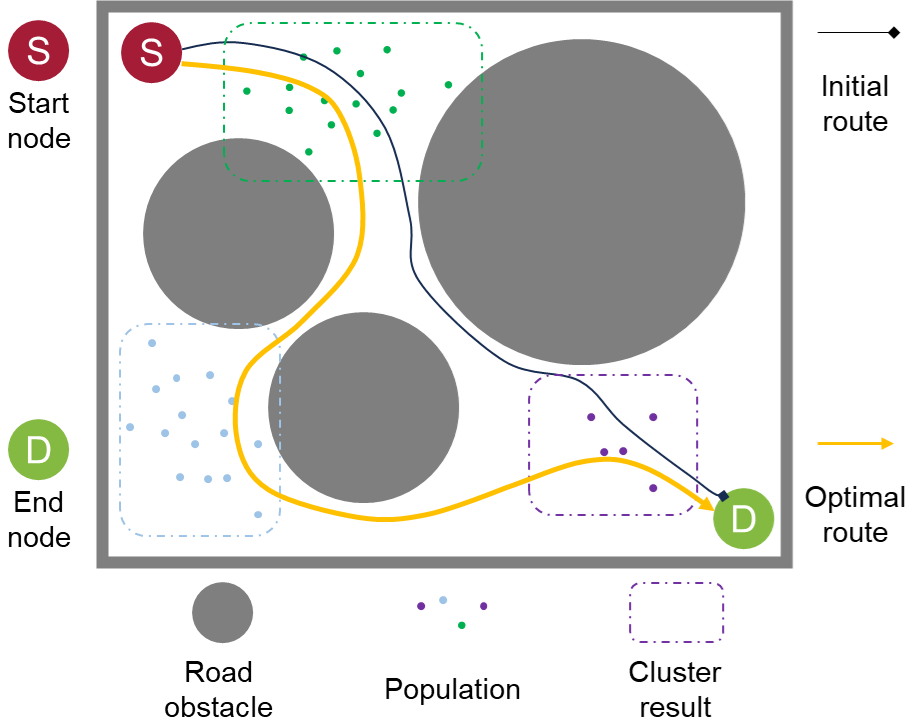}
		\caption{An example of clustering-based methods. Here clustering is done to provide user density (say, possible taxi customers or historically frequent driving routes). Then an optimal route incorporating user influence is redefined and can differ from the original one.}
		\label{fig:clusterbased}
\end{figure}

Clustering techniques are common useful tools for data dimension reduction and pattern recognition. In our route recommendation setting, clustering is usually used for finding similar route preferences among different users and/or for detecting similar road network structures and traffic flow patterns. By applying appropriate clustering techniques, the original problems can be solved structurally on a smaller scale, thus improving the search efficiency. 
An example is shown in Figure~\ref{fig:clusterbased}, which is to optimize driving routes based on historical user data.

Chen~\textit{et al.}~\cite{chenPersonalizedNavigationRoute2023} proposed a personalized navigation recommendation method based on user preferences. They first used a binary K-means algorithm to obtain the initial user preferences and then used differential perception to adaptively update preferences. Finally, they quantified the road network individually based on the user preferences and adopted the Tabu search algorithm to obtain the optimal path. 
Li~\textit{et al.}~\cite{liPhysicsguidedEnergyefficientPath2018} designed a physics-guided energy consumption model for path recommendation. They used K-means to aggregate trajectories to the closest scene and in turn, updated the signature of each scene based on the trajectories it includes.
Chen~\textit{et al.}~\cite{BPlanner2014} proposed a two-stage bidirectional method for nighttime bus route planning. The first stage was to gather densely populated hot spot areas and divide them into clusters. Then, they derived search rules to construct a bus route map based on the cluster-level information, thus achieving more efficient bus route planning.
Zhou~\textit{et al.}~\cite{SmartZhou2019} proposed a tourism route planning algorithm using a cluster-based iterative search. By using the tourist's residence as the clustering center, the algorithm could output a one-way shortest path from scenic spots to the clustering center. Then it generated a complete binary tree that outputs the optimal closed-loop route for tourism.
Lou and Gu~\cite{louAnalysis2022} improved the traditional $k$-means clustering analysis algorithm that relies less on the initial clustering centers and applied the improved method to the design of intelligent tourism route planning schemes.
Yu and Yang~\cite{Yuco2023} used hierarchical clustering to find similar routing needs and matched them with the most cost-efficient tourism routes.
Chen~\textit{et al.}~\cite{Chenper2021} analyzed the social relationships among users and obtained clusters in terms of friendship. Then, they compared the trajectories of people in the same cluster and recognized similar travel patterns. In this way, they reduced the search space by clustering and could recommend travel routes more efficiently.


\subsection{Constraint-based Methods}



\begin{figure}[h]
		\centering
		\includegraphics[width=0.9\linewidth]{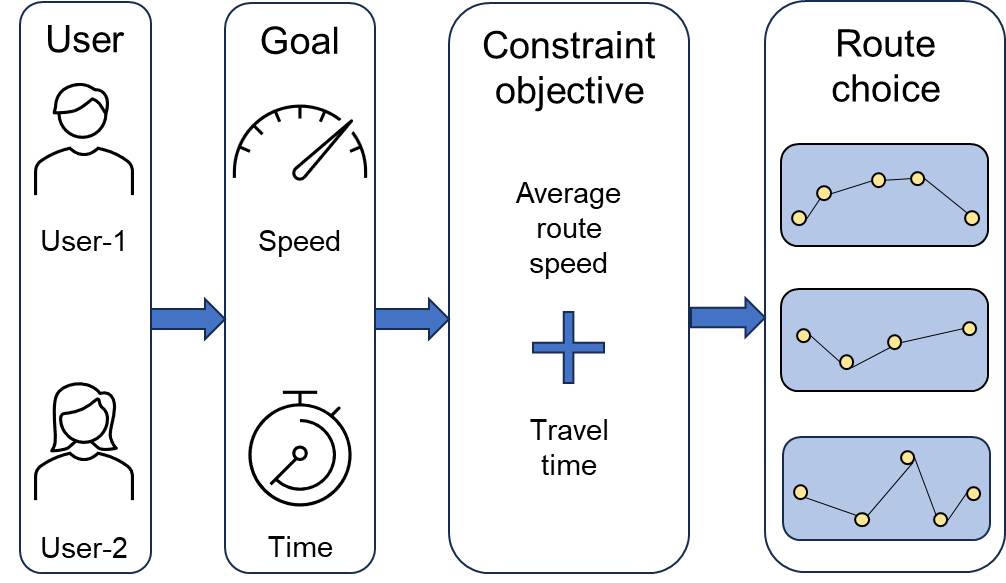}
		\caption{An example of constraint-based methods. Featured by establishing final optimization objectives based on user needs.}
		\label{fig:constraintbased}
\end{figure}

Constraint-based methods explicitly take into account the route requirements or preferences of users and model them as constraints in the optimization procedure.
An example is shown in Figure~\ref{fig:constraintbased}, which is to optimize route search based on user needs imposed as constraints.

Qu~\textit{et al.}~\cite{quCosteffectiveRecommenderSystem2014} designed a graphical representation of road networks based on historical taxi GPS trajectories and formulated an objective function with practical constraints to evaluate the profit of different driving routes. 
Salgado~\textit{et al.}~\cite{salgadoEfficientApproximationAlgorithm2018} dealt with an indoor route planning problem that is category-specific and multi-criteria. They explicitly required that each planned route must pass through at least one middle point in each given category, which was added as a constraint to the route search procedure.
Huang~\textit{et al.}~\cite{huangBackwardPathGrowth2015} proposed a dynamic programming method for mobile sequential recommendations for taxi drivers. Their method consists of a pre-processing stage and an online search stage where the former stage pre-computes potentially optimal sequences from a set of pickup points, and the latter stage selects the optimal driving route based on the pre-computed sequences. Such a two-stage optimization process was demonstrated more efficient.
Barth~\textit{et al.}~\cite{barthCYCLOPSCYCLeRoute2018} proposed a route recommendation system for cyclists, which incorporated three metrics (distance, height gain, suitability for cycling) and could recommend preferred routes to meet different cycling habits. 
Li~\textit{et al.}~\cite{liPhysicsguidedEnergyefficientPath2018} introduced a physics-guided energy consumption (PEC) model based on lower-order physical models, which estimates energy consumption as a function of vehicle parameters (such as mass and powertrain efficiency) and used this estimation for path selection in their proposed dynamic programming algorithm. 
Teng~\textit{et al.}~\cite{tengSemanticallyDiversePaths2021} solved a joint optimization problem for POI recommendation, which maximizes the diversity of POIs within a certain travel budget such as a distance/cost constraint.
Cheng~\textit{et al.}~\cite{chengTaxiCTaxiRoute2019} formulated taxi routing problems based on electric fields. They simulated taxis and passengers as electric charges and modeled their relationships as attraction or repulsion. Based on an urban transportation heat map, they proposed a taxi routing method that mimics charge motions. 
Reza~\textit{et al.}~\cite{rezaOptimalRouteStops2017} defined several optimization problems with constraints for carpooling routes and stop points planning. The goal of each problem is to minimize the cost and time of individual travelers under various constraints. 
Rajan and Ravishankar~\cite{rajanPhaseAbstractionEstimating2019} presented a new EV modeling method to find the optimal balance between travel time and energy consumption for electric vehicles. Their major contribution was to model energy consumption accurately at lower temporal granularity and then construct realistic vehicle speed profiles for real-world routes.
Lai~\textit{et al.}~\cite{laiUrbanTrafficCoulomb2019} applied Coulomb’s law to urban transportation when modeling the relationship between taxis and passengers and based on this physics model, designed a new route recommendation framework. 
Herschelman~\textit{et al.}~\cite{herschelmanConflictFreeEvacuationRoute2019} developed a method to generate evacuation routes with minimized evacuation time and the minimum number of movement conflicts.

\section{Deep learning for Route Recommendation}
\label{sec:Deep}

Early machine learning research primarily focused on logical reasoning, knowledge engineering, and simple models with theoretical guarantees. During the last decade, the rapid development of information infrastructures and technologies has allowed much larger-scale computation with big data.
Then, more portion of machine learning works are switching to data-driven learning paradigms with more complicated models, where the mainstream is deep learning.
Though neural networks have been researched for decades, they are gradually becoming feasible for solving practical problems lately. Compared to classic methods (Section \ref{sec:Traditional}), neural networks, especially deep models with delicate structural designs, have huge potential to discover very complex relationships among data. So in this section, we will see why and how deep learning methods prevail in route recommendation research.


\subsection{Why Deep Learning for Route Recommendation?}
The power of deep learning lies in its strong representation ability, realized by comprising multiple levels of non-linear data transformation layers~\cite{lecun2015deep}. 
With properly designed structures, deep learning models can extract very useful hidden knowledge behind data, either realistic knowledge or statistical correlations.
As for route recommendation tasks, richer information, either w.r.t data volumes or data types such as traffic/road information, trajectories, or user activities, allow deep models to learn novel and complex travel patterns of people. And consequently, more satisfactory and personalized routes can be designed for individuals.



There are three advantages of using deep learning for route recommendation.
First, deep models, consisting of many layers of flexible nonlinear transformations, can approximate very complex mapping between data domains and thus enjoy strong representation abilities. This allows us to process complicated real-world data for route recommendation.
Second, deep learning is more of a data-driven method. If with sufficient data, deep models can extract hidden knowledge behind data and thus may rely less on human knowledge (which might be biased and limited) and save labor. 
Third, many deep models can be trained in an end-to-end manner in route recommendation tasks, which enables the design of more complex and accurate models and their application in practice.


In the following, we will present deep route recommendation works and categorize them into five types: hybrid models, sequence-based models, graph neural network models, multi-modal models, and deep reinforcement learning models. Hybrid methods usually extend classic methods  (such as A*) with neural networks; sequence-based methods mainly model the temporal relations with recurrent neural networks (RNNs); graph neural networks (GNNs) further model spatio-temporal relations; multi-modal methods can process more complex data coming from different modalities; and lastly, deep reinforcement learning methods explicitly focus on making decisions for route recommendations.

DL models are often composed of different modules, for example, using GNNs to extract spatial features, RNNs for time series data, and MLP for feature fusion. As a result, there may be articles appearing in multiple categories. However, we do notice that their novelty contributed to certain parts of the model. So we categorize such articles according to their main contributions and emphasize their novelty under certain categories.

\subsection{Hybrid Approaches}
\label{subsec:hybrid}
\begin{figure}[h]
	\centering 
	\includegraphics[width=0.45\textwidth]{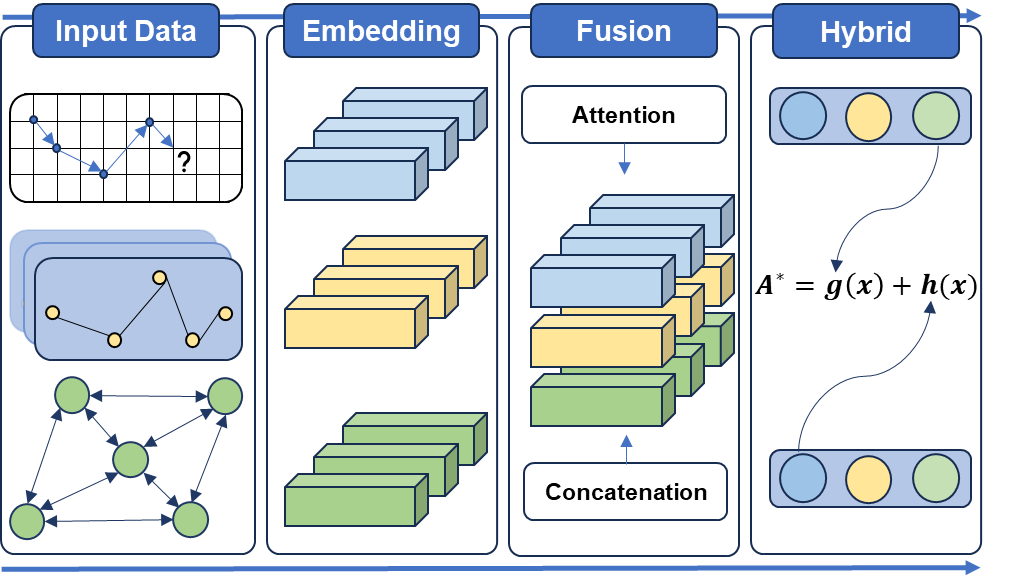}
	\caption{Hybrid Route Recommendation Approaches. Feature extraction through deep learning modules, then integrated into the A* framework.} 
	\label{fig:Hybrid}
\end{figure}

Traditional route recommendation methods are developed based on solid theoretical foundations but are also limited by simplistic assumptions and models. To this end, people attempted to follow the same rigorous logic but also enhance the modeling parts with more complex neural networks. Hence, hybrid methods enjoy both good explainability and empirical performance. Here, we mainly introduce A*-based models, which are good representatives and many works fall into this category. A high-level pipeline is shown in~\figureautorefname~\ref{fig:Hybrid}.


Recall that A* aims to model a cost function $F(x)=g(x)+h(x)$, where $g(x)$ is a base cost function for node $v_x$ and $h(x)$ is a heuristic function. The core of A* is to design a good $h(x)$ and estimate both $g(x)$ and $h(x)$ accurately.
For example, an early attempt was~\cite{wangEmpoweringSearchAlgorithms2019,wangPersonalizedRouteRecommendation2022}, which proposed using neural networks to automatically learn the cost functions of A* for personalized route recommendation tasks.
Wu \textit{et al.}~\cite{wuLearningEffectiveRoad2020} extended A* using neural networks to model complex traffic information and applied it to the recommendation of fast-traveling routes.
Yang \textit{et al.}\cite{yangNoahNeuraloptimizedSearch2021} proposed a new method named Noah, which combines A* with a graph neural network to calculate an approximate graph editing distance more efficiently and intelligently. 
Agostinelli \textit{et al.}~\cite{agostinelliObtainingApproximatelyAdmissible} proposed and studied batch versions of the A* algorithm. They applied approximate admissible transformations to the heuristic functions that are parameterized by deep neural networks. They showed that these heuristic functions could find optimal or bounded suboptimal solutions. 
Liu \textit{et al.}~\cite{liu2021ldferr} designed a fuel-efficient route recommendation model for long-distance driving based on historical trajectories. They combined neural networks with heuristic algorithms to automatically learn the optimal solutions of A*.
Lately, Wang \textit{et al.}~\cite{wangASNNFRRTrafficawareNeural2023} proposed ASNN-FRR, which extracts features from trajectory data with urban road information using AGCRN~\cite{bai2020adaptive} modules and estimates the values of $g(x)$ and $h(x)$ through a multi-layer perceptron (MLP) network. 
Besides A*, some works also adopted the classic Dijkstra algorithm in modern models.
Jain \textit{et al.}~\cite{jainNeuroMLRRobustReliable2021} proposed an inductive algorithm and a generative model called NEUROMLR for robust and reliable route recommendations. They convert the original computational task to the problem of finding the shortest path in a weighted graph and solve it by Dijkstra’s Algorithm, which has guaranteed optimality and accessibility.
Liu \textit{et al.}\cite{liuIntegratingDijkstraAlgorithm2020} utilized a deep inverse reinforcement learning (IRL) method to recommend delivery routes. They used the Dijkstra algorithm instead of value iteration to determine the current best strategy and calculate the gradients of the IRL model.
Other work such as Huang \textit{et al.}~\cite{huangMultiTaskTravelRoute2021} proposed a multi-task route recommendation framework in contrast to single-planning methods that only work for specific tasks. They used deep learning techniques to extract path representations and employed beam search algorithms to provide recommendations for multiple related tasks.

\subsection{Sequence-based Approaches}
\begin{figure}[h]
	\centering 
	\includegraphics[width=0.4\textwidth]{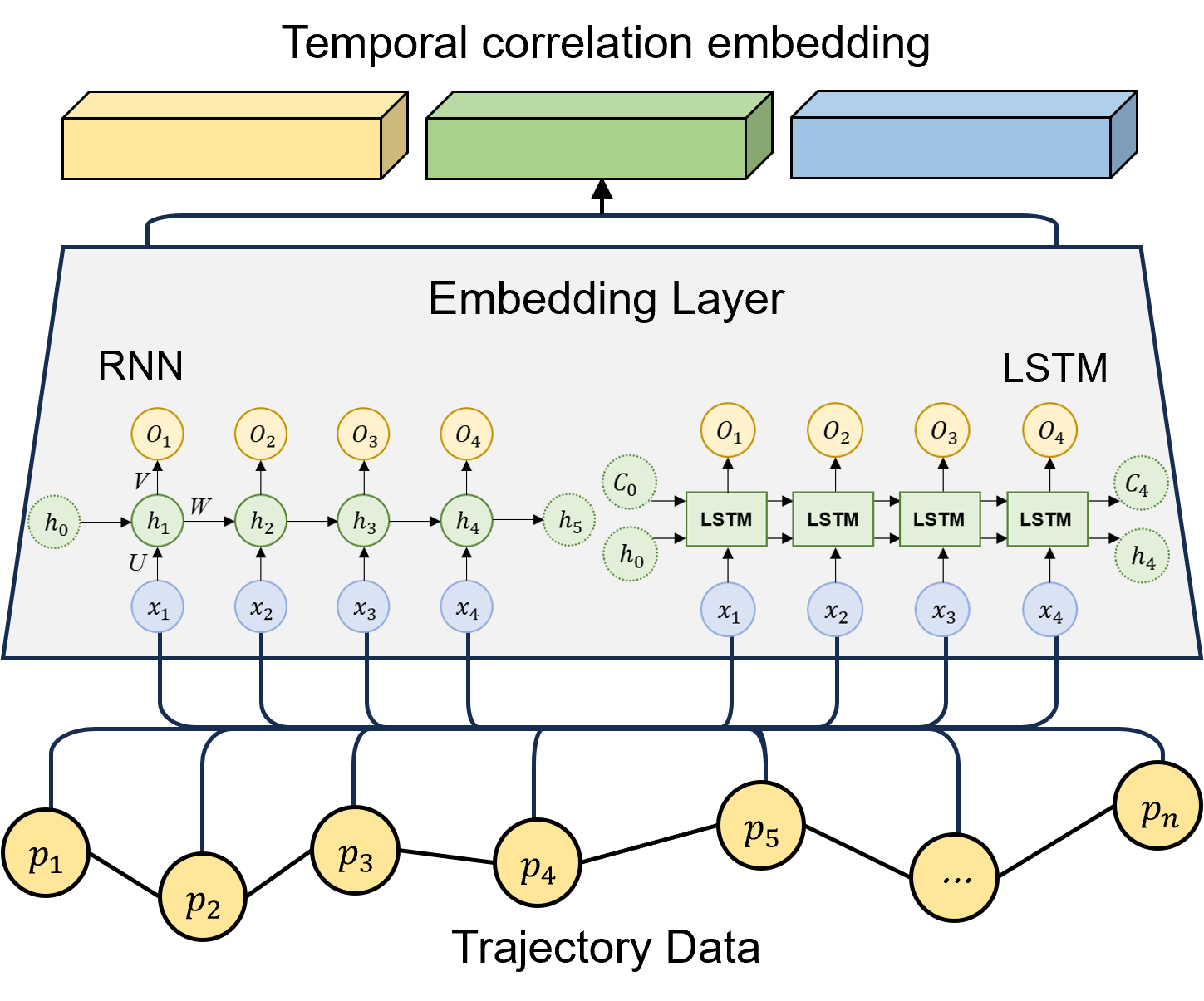}
	\caption{Sequence-based methods using RNNs for extraction of temporal relations in sequential data.} 
	\label{fig:RNN}
\end{figure}

Temporal relation is a key factor in route-related tasks, which can be modeled as time-series problems in general. Therefore, sequence-based models such as ARIMA, recurrent neural networks (RNNs) and their variances, and attention-based models like Transformers are now widely used in route recommendation methods. ~\figureautorefname~\ref{fig:RNN} shows a general example of using RNNs.


As a good representative, RNN~\cite{giles1994dynamic} is a classic autoregressive model that can capture the temporal correlations in sequential data. By summarizing the knowledge obtained from the previous data in a process, RNNs make predictions for the next timestamp based on the summarized information.
A typical calculation graph of an RNN model is shown in \equationautorefname~\eqref{eq:RNN}. 
\begin{equation}
\begin{aligned}
O_t &=g\left(V \cdot S_t\right),  \\
S_t &=f\left(U \cdot X_t+W \cdot S_{t-1}\right),
\end{aligned}
\label{eq:RNN}
\end{equation}
where $X_t, S_t, O_t$ respectively represent the observed input, summarized state, and output value at time $t$, $U, W, V$ respectively represent the input parameters, transfer parameters, and output parameters. Through this chain of calculation, one can see that the output $O_t$ is not only related to the current input $X_t$ but also affected by the information extracted from the previous moment $S_{t-1}$.
As vanilla RNNs can suffer from cataphoric forgetting and gradient varnish or explosion problems, Long Short-Term Memory (LSTM) networks~\cite{Hochreiter1997Long} and Gated Recurrent Unit (GRU) networks~\cite{chung2014empirical} were developed with additional memories to enhance longer-sequence modeling with more stable training processes.

The following examples are RNN-based methods for route recommendation. 
An early work~\cite{zhangWalkingDifferentPath2018} used RNN to extract feature representations of different POI nodes and added maximum difference constraints to recommend a path with diverse scenes for users.
Wang~\textit{et al.}~\cite{wangEmpoweringSearchAlgorithms2019,wangPersonalizedRouteRecommendation2022} proposed a personalized route recommendation method, which uses GRU segments to extract features from numerous user trajectories and employs the attention mechanism to increase the prediction accuracy. 
Similarly, Wen~\textit{et al.}~\cite{wenGraph2RouteDynamicSpatialTemporal2022} used GRUs to extract the temporal correlations in trajectory data, which helped to improve route recommendations for cargo deliveries.
Huang~\textit{et al.}~\cite{huangMultiTaskTravelRoute2021} proposed a multi-task prediction framework for route recommendation, which uses LSTM to fuse POI features, user features, and current path features, and applies a greedy network search algorithm to output results for multiple tasks. 
Hahn~\textit{et al.}~\cite{hahnPredictiveCollisionManagement2020} aimed to monitor collision information on roads and then recommend safer routes. They extracted features of collision accidents using LSTM and used them to predict the next step of moving positions. 
Liu~\textit{et al.}~\cite{liu2021ldferr} combined a bidirectional GRU and attention mechanism to estimate the fuel consumption in long-distance driving routes and recommended energy-efficient routes for drivers. 
Fu and Lee~\cite{fuProgRPGANProgressiveGAN2021} designed a progressive route recommendation method based on generative networks, which extracts knowledge from the selection of the next node using LSTM.
Wang~\textit{et al.}~\cite{wangASNNFRRTrafficawareNeural2023} introduced a travel route recommendation algorithm using AGCRN modules~\cite{bai2020adaptive}, which combines an RNN and a graph convolutional neural network (GCN) to capture spatial and temporal correlations in a fine-grained dimensional space. 

Besides RNNs which are autoregressive models, the Transformers~\cite{vaswani2017attention} are attention-based models that can process much longer sequences without suffering from the forgetting problem. 
The Transformer architecture consists of several key modules such as positional encoding, self-attention (Eq.~\eqref{eq:selfatt}), multi-head attention (Eq.~\eqref{eq:multiheadatt}), and feed-forward neural networks.


\begin{equation}
    \text{Attention}(Q,K,V) = \text{softmax}\left(\frac{QK^T}{\sqrt{d_k}}\right)V
    \label{eq:selfatt}
\end{equation}
Where matrices $Q$, $K$, and $V$ correspond to the query, key, and value components, respectively, and $d_k$ denotes the dimension of the key.

\begin{equation}
\begin{aligned}
\text{MultiHead}(Q,K,V) = \text{Concat}(head_1,...,head_h)W^O \\
\text{where } head_i = \text{Attention}(QW_i^Q, KW_i^K, VW_i^V),
\end{aligned}
\label{eq:multiheadatt}
\end{equation}
Where matrices $W_i^Q$, $ W_i^K$, $W_i^V$, and $W^O$ are learnable weight matrices within the model.
So we see that Transformers essentially differ from autoregressive models by using attention to process long sequences.


Therefore, many studies began to explore route recommendation methods based on Transformer.
For instance, Bhumika and Debasis Das~\cite{BhumikaMARRS} introduced a multi-task learning framework, which uses Transformers to extract spatial, temporal, and semantic correlations, and adopted epsilon constraint techniques to optimize routes that satisfy multiple objective functions.
Wang~\textit{et al.}~\cite{WangPersonalizedLong2022} studied long-distance travel trajectories, from which they extracted fuel-consumption factors using Transformer. Further combining a genetic optimization algorithm, they developed a model that could recommend more fuel-economic routes.
Wang~\textit{et al.}~\cite{WangPersonalizedSpecified2021} modeled route recommendation as a process of inferring a high-sampling rate fine trajectory from a low-sampling rate coarse trajectory composed of origin-destinations and way-points. Their proposed Seq2Seq model using multi-head self-attention mechanisms can automatically adjust the sampling weights to capture more accurate spatio-temporal correlation.
Wang~\textit{et al.}~\cite{WangQuery2Trip2023} proposed a new de-biased representation learning method named TripRec to implement the sequence generation method from Query to Trip. Based on the encoded representations, they further used Transformer for decoding and generated final routes.
Ngai Lam Ho and Kwan Hui Lim~\cite{HoPOIBERT2022} proposed POIBERT, which uses the BERT language model to recommend personalized itineraries based on users' preferences and their past POI selections.

\subsection{Graph-based Approaches}

\begin{figure}[h]
	\centering 
	\includegraphics[width=0.45\textwidth]{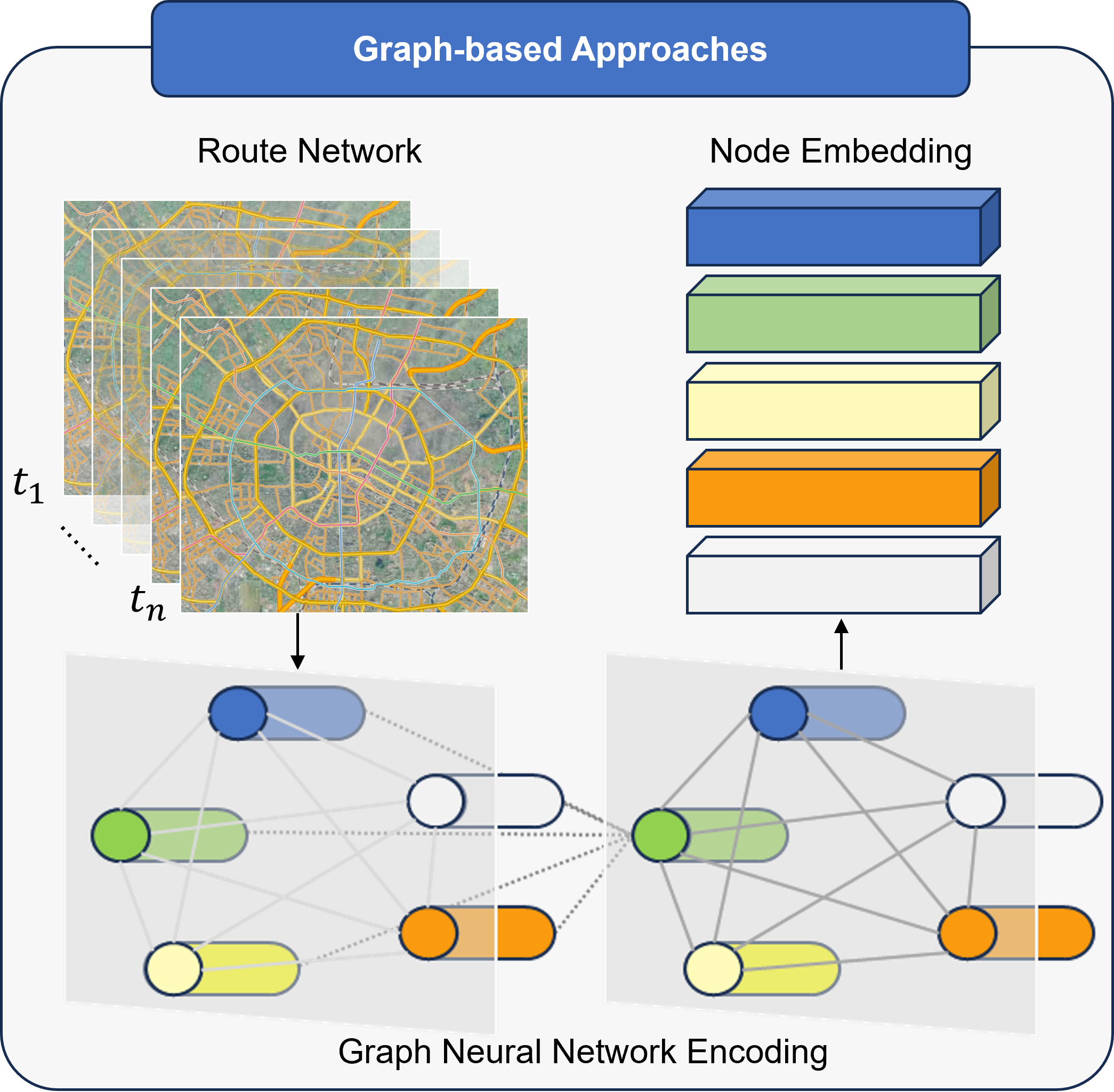}
	\caption{Graph based methods. A method for feature extraction of structured data using graph networks.} 
	\label{fig:graph}
\end{figure}

Besides temporal correlations in route-related data, spatial information is also a critical factor in route recommendation tasks. 
In urban situations, many data have strong structural characteristics and can be naturally modeled as temporal-spatial graphs, such as knowledge graphs and relational recommendations. Many urban transportation networks are irregular, large-scale, and expanding, which fails many classic methods models.
In response to such new challenges, graph neural networks (GNNs)~\cite{scarselli2008graph} recently have shown great potential in temporal-spatial research. GNNs are used to process graph-structured data which can broadly include numerous data types such as networks, texts, molecular biology, chemistry, and physics. In urban computing scenarios, routes, trajectories, and road information can be naturally modeled by GNNs with or without additional neural network modules (e.g. RNNs, CNNs, deep encoders, or attention-based modules). The general process is shown in~\figureautorefname~\ref{fig:graph}.

Currently, the main GNN types are Graph Convolutional Networks (GCNs, Equation~\eqref{eq:GNN}) and Graph Attention Networks (GATs, Equation~\eqref{eq:GAT}). Notably, GATs have employed the attention mechanism to weigh the influence of neighboring nodes, which endows GATs with enhanced flexibility and efficacy in processing various types of graphs.

\begin{equation}
H^{(l+1)} = \sigma\left( \tilde{D}^{-\frac{1}{2}} \tilde{A} \tilde{D}^{-\frac{1}{2}} H^{(l)} W^{(l)} \right)
\label{eq:GNN}
\end{equation}
Where $H^{(l)}$ denotes the feature matrix of nodes at the $l$-th layer, $W^{(l)}$ the weight matrix of that layer, $\tilde{A}$ the adjacency matrix with self-connections, $\tilde{D}$ the degree matrix of $\tilde{A}$, and $\sigma$ the nonlinear activation function.

\begin{equation}
\begin{aligned}
\alpha_{ij} &= \frac{\exp\left(\text{LeakyReLU}\left(\mathbf{a}^T \left[ W h_i \| W h_j \right]\right)\right)}{\sum_{k \in \mathcal{N}_i} \exp\left(\text{LeakyReLU}\left(\mathbf{a}^T \left[ W h_i \| W h_k \right]\right)\right)} \\
h'_i &= \sigma\left(\sum_{j \in \mathcal{N}_i} \alpha_{ij} W h_j\right),
\end{aligned}
\label{eq:GAT}
\end{equation}
Where $h_i$ and $h_j$ represent the feature vectors of nodes $i$ and $j$ respectively, $W$ the shared weight matrix, $\alpha_{ij}$ the attention weight of node $i$ w.r.t. its neighbor $j$, $\mathbf{a}$ the parameter of the attention mechanism, and $\sigma$ the activation function.



Many studies have applied GNNs to urban route recommendation tasks, which have considered various factors such as road traffic information, user preferences, historical trajectories, etc. 
For example, Wang~\textit{et al.}~\cite{wangASNNFRRTrafficawareNeural2023} used graph convolutional networks (GCNs) to represent traffic information and extract user preferences from historical trajectories for personalized route recommendations. 
Gao~\textit{et al.}~\cite{gaoDualgrainedHumanMobility2023} used graph a spatio-temporal network to fuse POI and track location information, and employed transfer learning techniques to address the data sparsity problem in travel route recommendation. 
Wang~\textit{et al.}~\cite{wangEmpoweringSearchAlgorithms2019,wangPersonalizedRouteRecommendation2022} used a GNN to estimate the road distance, travel time, and heuristic function $h(x)$ in A* based on the extracted features of the GNN. 
Wen~\textit{et al.}~\cite{wenGraph2RouteDynamicSpatialTemporal2022} employed a GCN to encode the checkpoints and trajectories along goods delivery paths and extracted the spatio-temporal dependencies for providing more profitable routes. 
Wu~\textit{et al.}~\cite{wuLearningImprovementHeuristics2022} addressed the traveling salesman problem by using graph attention networks for path node embedding and applied their model for delivery route recommendations. 
Wu~\textit{et al.}~\cite{wuLearningEffectivelyEstimate2019} utilized a graph attention network to extract features from road vehicle trajectory and speed information and improved upon the A* constraint method. 
Wu~\textit{et al.}~\cite{wuLearningEffectiveRoad2020} constructed a three-level graph neural network structure to learn the representation of different functional areas in cities and empirically verified the effectiveness of the additional features used in route recommendations. 
Jain~\textit{et al.}~\cite{jainNeuroMLRRobustReliable2021} developed NeuroMLR, which learns a generative model from historical trajectories by conditioning on three explanatory factors: the current location, the destination, and real-time traffic conditions. The conditional distributions are learned through a combination of Lipschitz embedding with a GCN using historical trajectory data.
Yang~\textit{et al.}~\cite{yangNoahNeuraloptimizedSearch2021} applied graph attention networks and pre-trained models to embed graph-structured data and combined A* to compute approximate graph edit distance in a more effective and intelligent way.

As we have seen, GNNs can address some challenges of data representation and information extraction problems in graph-structured data. Compared with sequence-based models that only extract temporal correlations, GNNs can capture both temporal and spatial dependencies in many urban computing tasks, where the models need to process various types of information such as user activities, trajectories, road information, traffic flows, and other physical factors. 

\subsection{Multi-modal Approaches}
\label{subsec:multi-modal}
\begin{figure}[h]
		\centering
		\includegraphics[width=0.8\linewidth]{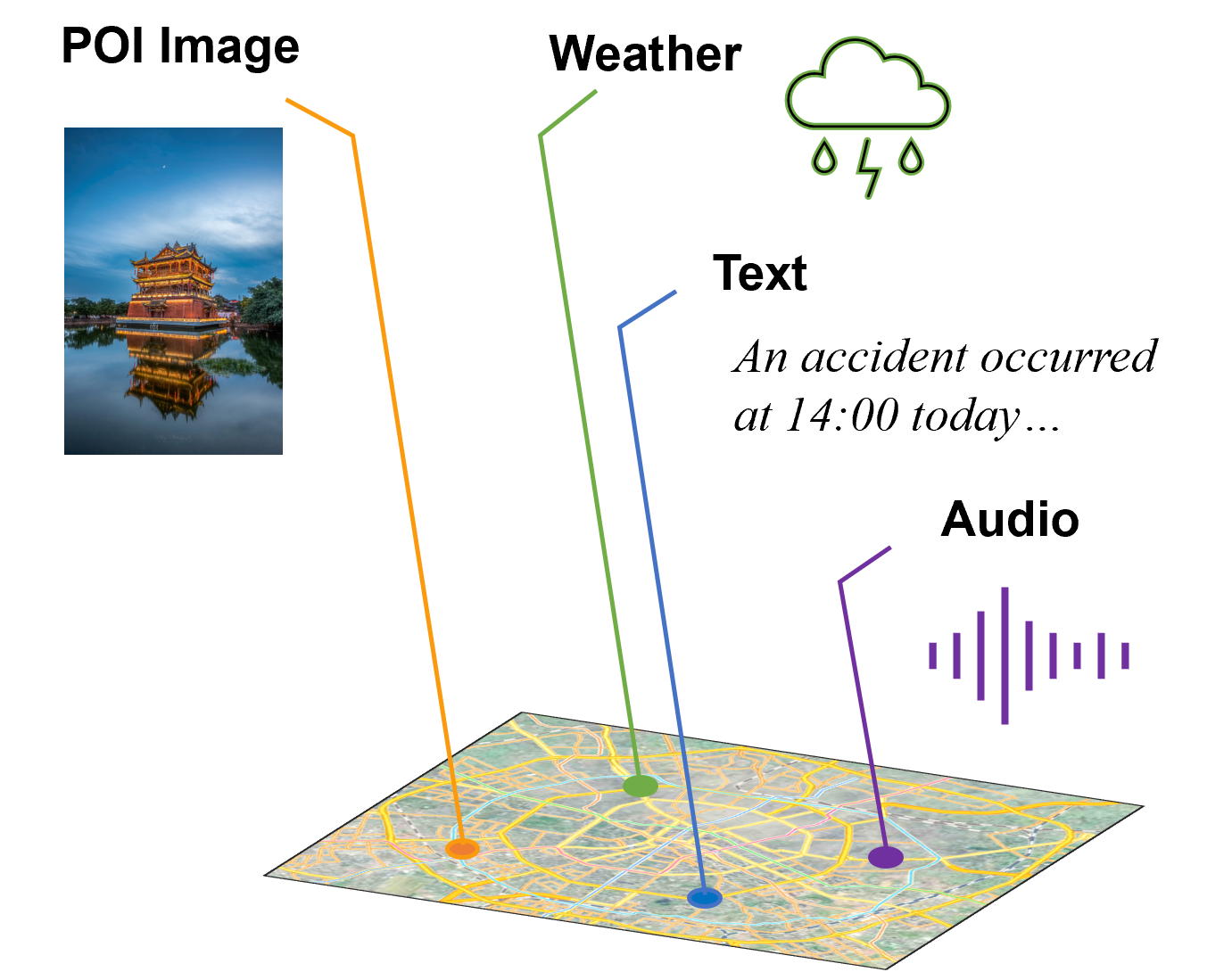}
		\caption{\textbf{Input modalities: POI images, texts, audio, and weather}. This figure illustrates the synergistic use of diverse data types to enrich route recommendations by incorporating real-time contextual information along with physical transportation networks.}
		\label{chutian1}
\end{figure}

\begin{figure}[h]
		\centering
		\includegraphics[width=0.9\linewidth]{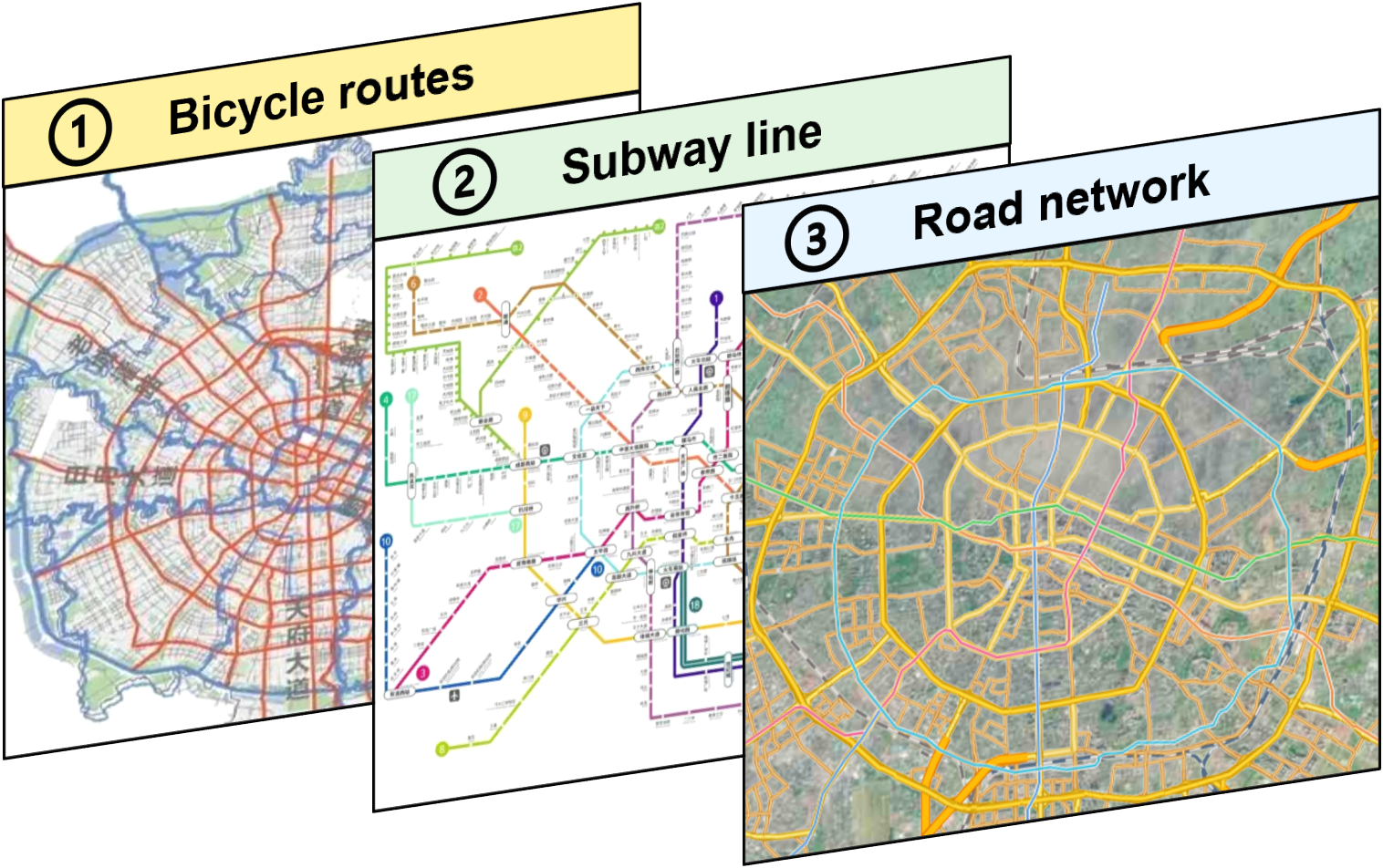}
		\caption{\textbf{Input modalities: bicycle routes, subway lines, and vehicle road networks}. \ding{192}, \ding{193}, and \ding{194} represent different transportation networks, which can be incorporated to enrich route recommendation tasks.}
		\label{chutian2}
\end{figure}

Most of the existing route recommendation methods process two types of data - trajectories and road network information. However, with the development of urban information systems, many more types of data are recorded, and thus single-modal methods may not be sufficient to meet the travel needs of users.
Meanwhile, multi-modal learning paradigms have been shown very effective in other research areas such as computer vision~\cite{WANG2024102132} and natural language processing~\cite{DBLP:journals/taslp/WangYLZL23,cao2022image}. 
Therefore, multi-modal route recommendation methods are a promising research area. New modalities in urban computing include but are not limited to weather information, different types of transportation networks, scenic images, texts, and audio.
In this sub-section, we review the literature from two perspectives: one mainly fuses images, texts, audio, and weather data (Figure~\ref{chutian1}), and the other fuses different types of transportation data (Figure~\ref{chutian2}).

Point-of-interest (POI) images and text information, when fused with historical trajectory data, have the potential to improve the route recommendation results with richer scenic information. 
For example, Zhang~\textit{et al.}~\cite{zhangWalkingDifferentPath2018} developed a system that recommends routes with diversified scenes. The diversity was measured by city views, which come from publicly available data such as Google Street View images. 
Bi~\textit{et al.}~\cite{biSpeakNavVoicebasedNavigation2021} developed SpearkNav, a navigation system that allows users to describe routes through voice and supports clue-based route retrieval. This system includes a language-understanding module and a route-designing module based on the voice input. 
Park~\textit{et al.}~\cite{parkStudyTopicModels2022} crawled comments from the Landstar website and used Tagxedo, a text mining and network analysis tool to design travel routes for tourists in South Korea. 
Padia~\textit{et al.}~\cite{padia2019sentiment} analyzed user interests based on the sentiment behind their textual comments of visited locations and then recommended travel plans that can match their interests. 
Campigotto~\textit{et al.}~\cite{campigottoPersonalizedSituationAwareMultimodal2017} proposed to learn user preferences through a Bayesian approach. They posed formatted questions to obtain initial user information and encoded it as the prior knowledge of their Bayesian model.
Hu~\textit{et al.}~\cite{huAGREEAttentionBasedTour2019} utilized attention mechanism to dynamically estimate the influence of personal data. They used an attention network to focus on influential data in multi-modal POI sequences and then adopted bi-directional recurrent units (Bi-GRU) to generate recommended routes.

In reality, people's travel plans can be chosen from different modes of transportation or a mixture. So, recommendation algorithms in certain scenarios need to consider such multiple choices and find the best way (or a mix) of transportation.
For instance, Georgakis~\textit{et al.}~\cite{georgakisMultiModalRoutePlanning2019} proposed a multi-modal path planner that allows different transportation modes for different parts of the travel plan.
Bucher~\textit{et al.}~\cite{bucherHeuristicMultimodalRoute2017} developed an algorithm based on user profiles and generated personalized multi-modal routes among a set of feasible travel options.
Herzog~\textit{et al.}~\cite{herzogRouteMeMobileRecommender2017} proposed RouteMe, a mobile recommendation system for personalized multi-modal routes. They combined Collaborative Filtering (CF) with knowledge-based recommendations to enhance the quality of route recommendations.
Cheng~\textit{et al.}~\cite{chengTERPTimeEventDependentRoute2019} tackled the aggregation problem of transportation modes by modeling travel time, arrival time, bicycle ownership, and transfer time among different modes of transportation. They introduced TERP, a path planner that considers time events and optimizes both travel time and reliability.
Zhang and Arentze~\cite{zhangPersonalizedMultimodalRoute2014} estimated the distribution of people's different travel patterns and modes. Based on this, they recommended personalized travel choices to users that are similar to each other.
Yu and Lu~\cite{yuAdvancedMultimodalRouting2012} developed an approach that automatically constructs a pedestrian network after analyzing both the pedestrian infrastructure and semantic analysis.
Liu~\textit{et al.}~\cite{liuJointRepresentationLearning2019} extracted multi-modal transportation maps from large-scale map query data. Based on this, they inferred user correlations and origin-destination correlations, and used such information to recommend online multi-modal routes. In their subsequent work~\cite{liuHydraPersonalizedContextAware2019,liu2020multi,liuIncorporatingMultiSourceUrban2022,liuUnifiedRouteRepresentation2023b}, they further incorporated a pre-training mechanism and a contrastive learning approach to analyze trajectories and road networks.

\subsection{Deep Reinforcement Learning Related Approaches}
\begin{figure}[h]
	\centering 
	\includegraphics[width=0.45\textwidth]{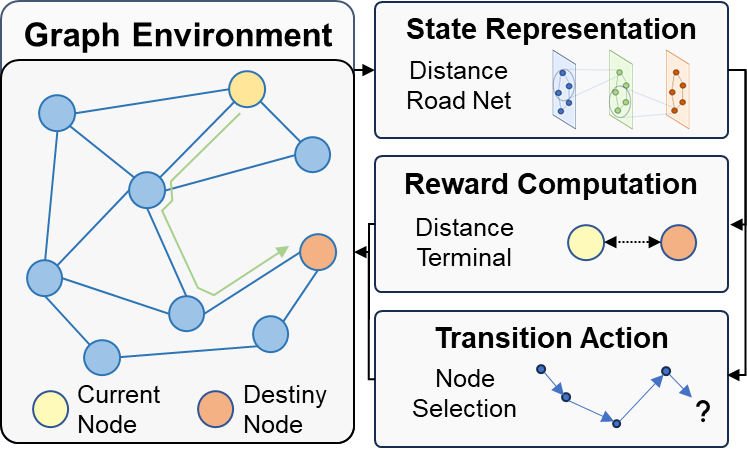}
	\caption{Deep Reinforcement Learning Approaches. } 
	\label{fig:Reinforcement}
\end{figure}

Reinforcement Learning (RL) is a learning paradigm involving an agent interacting within an environment strategically to maximize the rewards it can receive. The agent perceives the environment's states and acts accordingly. So the key of RL is to learn a good policy, or equivalently, a Q function $Q(s, a)$ for the agent. 
While classic RL learns relatively simple Q functions or policies, modern RL relies on deep neural nets to handle more complex states, actions, and policies, allowing agents to make decisions in more intricate tasks and environments. Therefore, Deep Reinforcement Learning (DRL) has demonstrated more power in various research areas, including route planning.
The general elements of DRL for route recommendation are shown in ~\figureautorefname~\ref{fig:Reinforcement}.


Many DRL methods are applied to urban route recommendations, which have considered various factors such as cost, distance, evacuation, driving preferences, and spatio-temporal patterns. 
For example, Xia~\textit{et al.}~\cite{xiaEfficientNavigationConstrained2022} studied the classic constrained shortest paths (CSP) problem and developed an iterative search process. They designed an adaptive controller based on RL to effectively control the expansions of candidate options in each search iteration.
Chen~\textit{et al.}~\cite{chenCuRLGenericFramework2023} proposed a generic bi-criteria optimal path-finding framework (cuRL) based on DRL, which co-optimizes two objective functions - minimized distance and optimized extra cost incurred when satisfying user's preferences. 
Bi~\textit{et al.}~\cite{biEvacuationRouteRecommendation2019} proposed an evacuation route recommendation method, which first uses an autoencoder to perform dimensional reduction and then employs an RL-based route-selection algorithm to design routes. 
Liu~\textit{et al.}~\cite{liuIntegratingDijkstraAlgorithm2020} developed a deep inverse RL (IRL) algorithm to learn the driving preferences of transportation personnel from their GPS trajectories and then recommend their preferred routes. 
Liu and Jiang~\cite{liuPersonalizedRouteRecommendation2022} further improved the IRL method for taxi routing problems by considering real-time traffic conditions.
Agostinelli~\textit{et al.}~\cite{agostinelliObtainingApproximatelyAdmissible} adopted RL to learn admissible heuristics in the A* search to solve path-finding problems that have large state spaces. 
Wu~\textit{et al.}~\cite{wuLearningImprovementHeuristics2022} proposed a DRL framework to automatically learn search heuristics, which are frequently handcrafted, for routing problems. They applied their learned heuristics to the traveling salesman problem (TSP) and the capacitated vehicle routing problem (CVRP), shown with superior performance.
Ji~\textit{et al.}~\cite{jispatiotemporalFeatureFusion2020} modeled a taxi routing problem as a decision process and designed an adaptive DRL method, which can better integrate spatio-temporal features and recommend more effective routes.

\begin{table*}[h]
\renewcommand{\arraystretch}{1.5}
\caption{Applications of route recommendations.}
	\centering
    \small
	\begin{tabularx}{\textwidth}{lXXX}
		\toprule
		\textbf{Application Category}&  \textbf{Different Aspects} & \textbf{References}  \\
		\midrule
		
\multirow{3}{*}{\textbf{Tourism Route Recommendation}} 
&Attraction-based& \cite{parkStudyTopicModels2022}~\cite{gaoAdversarialHumanTrajectory2023}~\cite{ahmadiBestCompromiseInRouteNearest2017}~\cite{chengTravelRouteRecommendation2021}~\cite{weiConstructingPopularRoutes2012}~\cite{suCrowdPlannerCrowdbasedRoute2014}~\cite{chenLearningPointsRoutes2016}
\cite{duMakeTravelHealthier2019}~\cite{zhengNovelMultiObjectiveMultiConstraint2021}~\cite{taylorTravelItineraryRecommendations2018}\\
&  Travel-time-based     & \cite{liuRealtimePersonalizedRoute2014}~\cite{xuTravelRouteRecommendation2016}~\cite{fitzgeraldOnlineRouteReplanning2021}\\
&Navigation system development&\cite{suCrowdPlannerCrowdbasedRoute2014}~\cite{friggstadOrienteeringAlgorithmsGenerating2018} \\ \cline{2-3}

 \multirow{6}{*}{\textbf{Economic Route Recommendation}} &  Revenues   &\cite{quCosteffectiveRecommenderSystem2014}~\cite{quProfitableTaxiTravel2020}~\cite{qianSmartRecommendationMining2012}~\cite{jispatiotemporalFeatureFusion2020}~\cite{chengTaxiCTaxiRoute2019}~\cite{laiUrbanTrafficCoulomb2019}~\cite{huangBackwardPathGrowth2015}\cite{chenPersonalizedNavigationRoute2023}
 ~\cite{liu2021ldferr} \\
 &Fuel consumption& \cite{nguyenACObasedTrafficRouting2023}~\cite{geEnergyefficientMobileRecommender2010}~\cite{liPhysicsguidedEnergyefficientPath2018}~\cite{liu2021ldferr} \\
 &Electric consumption& \cite{gareauEfficientElectricVehicle2019}~\cite{rajanPhaseAbstractionEstimating2019}~\cite{deoliveiraMinMaxRoutingCase2017} \\
 &Crowdsourcing solution design&\cite{wenGraph2RouteDynamicSpatialTemporal2022}~\cite{costaInrouteTaskSelection2018}~\cite{liuIntegratingDijkstraAlgorithm2020}~\cite{wuLearningImprovementHeuristics2022}~\cite{sunOnlineDeliveryRoute2019}~\cite{costaOnlineInRouteTask2020}~\cite{chenOnlineRoutePlanning2021} ~\cite{papadopoulosPersonalizedFreightRoute2023}\\
     &Group mobility plan &\cite{rezaOptimalRouteStops2017}~\cite{hashem2013group}\\ \cline{2-3}
 
\multirow{4}{*}{\textbf{Personalized Route Recommendation}}   
& Time-oriented &\cite{sacharidisFindingMostPreferred2017}~\cite{zhuFineRoutePersonalizedTimeAware2017}~\cite{deoliveiraesilvaPersonalizedRouteRecommendation2022}~\cite{huangTimedelayNeuralNetwork2017}~\cite{wangASNNFRRTrafficawareNeural2023} \\
& Distance-oriented
&\cite{xiaEfficientNavigationConstrained2022}~\cite{wuLearningEffectiveRoad2020}~\cite{kimProcessingTimedependentShortest2014} \\ 
& Preference-oriented
&\cite{liuPersonalizedRouteRecommendation2022}~\cite{daiPersonalizedRouteRecommendation2015}~\cite{wangPersonalizedRouteRecommendation2022}~\cite{tengSemanticallyDiversePaths2021}~\cite{zhangWalkingDifferentPath2018}~\cite{linGoalPrioritizedAlgorithmAdditional2020}\\
& Difference-oriented &\cite{chenEffectiveEfficientReuse2019}~\cite{chondrogiannisFindingKDissimilarPaths2018}~\cite{hackerMostDiverseNearShortest2021}\\
\cline{2-3}

    \multirow{2}{*}{\textbf{Security-related Route Recommendation}}    & Privacy protection   & \cite{islamPrivacyEnhancedPersonalizedSafe2021}~\cite{hashem2018crowd}~\cite{GALBRUN2016160}\\
    & Safe evacuation& \cite{mohdnordinApplicationAlgorithmAmbulance2011}~\cite{herschelmanConflictFreeEvacuationRoute2019}~\cite{biEvacuationRouteRecommendation2019}~\cite{khanMacroServRouteRecommendation2017}\\ \cline{2-3}
    
       \multirow{1}{*}{\textbf{Indoor Route Recommendation }}  &   /  & \cite{salgadoEfficientApproximationAlgorithm2018}~\cite{liuIKAROSIndoorKeywordAware2022}~\cite{liBarrierFreePedestrianRouting2021}\\
		\bottomrule
	\end{tabularx}%
	\label{tab:Applications}%
\end{table*}%

\section{Applications}
\label{sec:Application}
Applications related to route recommendation are rich. Here we mainly present the latest applications in the current research community, whereas the classic ones (like the shortest-path finding problem) are already well-known and thus omitted.
\tableautorefname~\ref{tab:Applications} gives an overview.

\subsection{Tourism Route Recommendation}
The key to tourism route recommendation is to match user's interests with different attractions on a potential route. Also, common basic factors such as travel time and cost need to be considered as well.
A large body of works focused on user interests~\cite{parkStudyTopicModels2022,gaoAdversarialHumanTrajectory2023,ahmadiBestCompromiseInRouteNearest2017,chengTravelRouteRecommendation2021,suCrowdPlannerCrowdbasedRoute2014,chenLearningPointsRoutes2016,huangMultiTaskTravelRoute2021,duMakeTravelHealthier2019,taylorTravelItineraryRecommendations2018}. They typically start with extracting user's travel interests from their historical trajectories and then recommend routes of which the POIs can best match user's interests. The matching process uses some similarity or consistency metrics. 
For example, Park and Liu~\cite{parkStudyTopicModels2022} applied the latent Dirichlet allocation (LDA) and Word2Vec models to match user interests with scenic spots. 
Gao~\textit{et al.}~\cite{gaoAdversarialHumanTrajectory2023} designed an adversarial network consisting of a generator and a reviewer (i.e. discriminator) and used the reviewer to distinguish between a route representation and a user's query representation. 
Huang~\textit{et al.}~\cite{huangMultiTaskTravelRoute2021} constructed a heterogeneous network based on the relationship between users and POIs, and used a network embedding method to learn the features of users and POIs. 
Zheng~\textit{et al.}~\cite{zhengNovelMultiObjectiveMultiConstraint2021} proposed a variable neighborhood search algorithm with a hybrid particle swarm genetic optimization algorithm to recommend Top-K travel routes. 
Some other works still considered travel time as the main factor of routes~\cite{liuRealtimePersonalizedRoute2014,xuTravelRouteRecommendation2016,fitzgeraldOnlineRouteReplanning2021}; and some aimed to design navigation systems for traveling routes~\cite{suCrowdPlannerCrowdbasedRoute2014,friggstadOrienteeringAlgorithmsGenerating2018}.

\subsection{Economic Route Recommendation}
%
Economic route recommendation mainly aims to design economic routes, i.e. minimizing some costs while also satisfying the user’s needs. It is a crucial task that affects energy consumption and revenues of urban transportation, which are key indicators of city life. There are different aspects to define costs, such as taxi revenue, fuel consumption, electric consumption, crowdsourcing solution design, and group mobility plan.


On taxi revenues, many works focused on how to construct an objective function that maximizes driver's income and design different optimization strategies~\cite{quCosteffectiveRecommenderSystem2014,quProfitableTaxiTravel2020,qianSmartRecommendationMining2012,jispatiotemporalFeatureFusion2020,chengTaxiCTaxiRoute2019,laiUrbanTrafficCoulomb2019,huangBackwardPathGrowth2015,chenPersonalizedNavigationRoute2023,liu2021ldferr}.
For example, Qu~\textit{et al.}~\cite{quCosteffectiveRecommenderSystem2014} developed a cost-effective recommending system for economic driving routes. 
Ge~\textit{et al.}~\cite{geEnergyefficientMobileRecommender2010} designed a mobile recommendation system that can recommend a range of parking locations for taxi drivers. 
Papadopoulos~\textit{et al.}~\cite{papadopoulosPersonalizedFreightRoute2023} proposed a coordinated pricing and routing scheme for a truck routing problem, which not only considers truck drivers' revenues but also improves the overall traffic condition in the network.

In contrast to revenue controls, many works directly study how to reduce fuel consumption~ \cite{nguyenACObasedTrafficRouting2023,geEnergyefficientMobileRecommender2010,liPhysicsguidedEnergyefficientPath2018,liu2021ldferr}. 
For instance, Li~\textit{et al.}~\cite{liPhysicsguidedEnergyefficientPath2018} developed a Physics-guided Energy Consumption model for solving the energy-efficient path selection problem which aims to find the path with the least expected energy consumption. 
Nguyen and Jung~\cite{nguyenACObasedTrafficRouting2023} introduced a decentralized traffic path system based on an ant colony optimization algorithm. Their system was shown to significantly reduce travel time and fuel consumption compared to existing systems. 
In addition, due to the development of electric vehicles (EVs), route recommendation for EVS poses a unique challenge that we need to consider the dynamic relationship between the charging locations and real-time power consumption~\cite{gareauEfficientElectricVehicle2019,rajanPhaseAbstractionEstimating2019,deoliveiraMinMaxRoutingCase2017}.
One representative work was by Gareau~\textit{et al.}~\cite{gareauEfficientElectricVehicle2019} that focused on route planning for EVs. They aimed to find the shortest path between two given locations while passing by several charging stations. The total time has considered the total driving time, the charging time, and the waiting time.

Crowdsourcing tasks also require efficient and economic distribution of routes, so many researchers have designed different solutions to meet the needs~\cite{wenGraph2RouteDynamicSpatialTemporal2022,costaInrouteTaskSelection2018,liuIntegratingDijkstraAlgorithm2020,wuLearningImprovementHeuristics2022,sunOnlineDeliveryRoute2019,costaOnlineInRouteTask2020,chenOnlineRoutePlanning2021,papadopoulosPersonalizedFreightRoute2023}. 
Costa~\textit{et al.}~\cite{costaInrouteTaskSelection2018} proposed an exact (but expensive) solution and other practical heuristic solutions for the multi-task route selection problem of crowdsourcing tasks.
Liu~\textit{et al.}~\cite{liuIntegratingDijkstraAlgorithm2020} applied a deep IRL algorithm to capture the preferences of traveling salesmen from their historical GPS tracks and recommend preferred routes, thus improving the delivery efficiency.

Lastly, traveling collectively in a group, such as ride-sharing, can largely reduce individual costs. There are works~\cite{hashem2013group,rezaOptimalRouteStops2017} that aimed to design a suitable route with multiple stop points for users traveling or commuting together. Since the goal is to minimize the total cost of individuals and also the ride-provider, they modeled this as a multi-objective optimization problem to solve.

\subsection{Personalized Route Recommendation}
%
Personalized route recommendation aims to provide users with travel suggestions that match their preferences. By analyzing the temporal and spatial patterns of user trajectories, personalized route recommendation methods can capture the user’s travel intentions or constraints and satisfy their needs. Compared with tourism route recommendations, personalized recommendations focus more on daily commutes. Common factors to be considered are time, distance, preference, and diversity.

Travel distance and time are still the primary considerations of users~\cite{sacharidisFindingMostPreferred2017,zhuFineRoutePersonalizedTimeAware2017,deoliveiraesilvaPersonalizedRouteRecommendation2022,huangTimedelayNeuralNetwork2017,wangASNNFRRTrafficawareNeural2023,xiaEfficientNavigationConstrained2022,wuLearningEffectiveRoad2020,kimProcessingTimedependentShortest2014}, while other works also take into account other preferences such as fuel consumption, visual scenes, driving habits, and diversity ~\cite{liuPersonalizedRouteRecommendation2022,daiPersonalizedRouteRecommendation2015,wangPersonalizedRouteRecommendation2022,tengSemanticallyDiversePaths2021,zhangWalkingDifferentPath2018,linGoalPrioritizedAlgorithmAdditional2020}.
Typically, they incorporated distance, time, and preferences as constraints into their learning objectives which also fuse user-based features (extracted from trajectories) and road traffic information.
For example, Zhu~\textit{et al.}~\cite{zhuFineRoutePersonalizedTimeAware2017} introduced the FineRoute algorithm, a personalized and time-sensitive route recommendation system. They considered three factors: user preferences, appropriate access time, and transition time in route generation, and measured the quality of a route at the appropriate travel time.


On the other hand, while most works recommend routes with similar patterns based on a user's history, other users may prefer to discover untraveled routes. Therefore, research work based on difference~\cite{chenEffectiveEfficientReuse2019,chondrogiannisFindingKDissimilarPaths2018,hackerMostDiverseNearShortest2021} is also a branch of personalized route recommendation. 
Specifically, Chen~\textit{et al.}~\cite{chenEffectiveEfficientReuse2019} proposed a parallel splitting and merging method (RSL-Psc), which constructs new routes by dividing and combining candidate sub-routes. The subtasks are executed in parallel to provide users with multiple different route choices.

\subsection{Security-related Route Recommendation}


%
Here we discuss two different aspects of security. One is to protect user privacy or other security issues in route recommendations.
In many route-based applications, user data are frequently uploaded to clouds. To protect data and user privacy, access control, encryption algorithms, and cryptographic methods are all important aspects.
Therefore, \cite{islamPrivacyEnhancedPersonalizedSafe2021,hashem2018crowd,GALBRUN2016160} proposed different ways to protect user privacy when using their data for route analysis based on historical data.
The other is to provide safe evacuation routes during emergencies.
For example, Mohdnordin~\textit{et al.}~\cite{mohdnordinApplicationAlgorithmAmbulance2011} developed an ambulance routing system that aims to ensure the availability of ambulances for emergencies within a limited time. 
Herschelman~\textit{et al.}~\cite{herschelmanConflictFreeEvacuationRoute2019} proposed a new method to minimize both the evacuation time and the number of mobile conflicts. 
Bi~\textit{et al.}~\cite{biEvacuationRouteRecommendation2019} proposed a prediction model based on the Markov decision process and aimed to design global optimal evacuation routes.
Khan~\textit{et al.}~\cite{khanMacroServRouteRecommendation2017} provided a scalable emergency evacuation service that can guide evacuees to a safe and least congested route.

\subsection{Indoor Route Recommendation}
%
While most of the existing work focuses on outdoor situations, indoor route recommendation, navigation, or evaluation is also important in the urban environment, especially in large malls, buildings, or complexes.
For example, Salgado~\textit{et al.}~\cite{salgadoEfficientApproximationAlgorithm2018} studied indoor route recommendations and considered the uniqueness of different indoor spots. They proposed to use Category-Aware Multi-criteria (CAM) route planning queries to categorize the spots and recommended routes with minimized distance and other costs.
Li~\textit{et al.}~\cite{liBarrierFreePedestrianRouting2021} studied a pedestrian route planning problem when the routes connect both indoor and outdoor scenes. They addressed three problems - route planning in indoor fine-grained scenes, indoor-outdoor route connection, and user query transfer in different scenes. 
Liu~\textit{et al.}~\cite{liuIKAROSIndoorKeywordAware2022} demonstrated an Indoor Keyword-aware Routing System (IKAROS) that can answer Top-K indoor keyword-aware routing queries. Through the queries, users can customize their indoor routes, and the system can visualize the designed routes for users.

\section{Conclusions and Future Directions}\label{sec:Conclusion}

Route recommendation now plays an increasingly important role in developing intelligent transportation and smart cities. We have seen various applications being extensively researched and deployed, such as public vehicle routing, traffic management, and safe evacuation. Some works are inherited from the classic methodologies (like Dijkstra's algorithm, A*) and extend them to modern situations. This line of methods has gradually evolved from using exact solutions or simple machine learning solutions to deep learning models that can deal with more complicated and larger-scale route recommendation problems. On the other hand, many more works attempt to explore new applications, cope with new problems, process new data modes, and develop new models accordingly. With richer data volumes and types, route recommendation problems can be solved on a fine-grained level, satisfying more specific user needs and preferences.

According to our experience, route recommendation research is still a rising topic and there are many open problems to be addressed. Meanwhile, since it is application-driven, there can be many new applications and new problems to explore. Below we list out a few of what we think are important and interesting research directions.

\begin{itemize}
    \item[$\bullet$] \textbf{Explainability.} It is widely criticized that deep learning suffers from poor explainability, though it is often considered the first choice for route recommendation research. Good explainability of models and interpretable results are crucial for route designers and city managers who aren't even machine learning practitioners~\cite{zhang2018visual}.
    We think the key to mitigating this problem is to understand the mechanisms of people's travel patterns in depth and inject such knowledge into modeling. A trend is still to combine the classic approaches (such as exact solutions and statistical analysis which enjoy good explainability and theoretical guarantees) and deep learning models which can handle more complex problems. We have seen a lot of such practices in Section~\ref{subsec:search}, \ref{subsec:prob} and \ref{subsec:hybrid}. Other general solutions for improving the interpretability of deep models are to do model visualization or to interpret black-box models through their output results~\cite{chenSurveyTrafficData2015a}. 
    In addition, demonstrating the strong generalization of models through transferability is also an interesting topic~\cite{Yosinski2014}. For route recommendation, many tasks need to deal with multi-domain or multi-modal data, which necessitates the research of model transfer and generalization ability across different domains.
    
    \item[$\bullet$] \textbf{Multi-modal learning.} Though we have mentioned some multi-modal route recommendation work in Section~\ref{subsec:multi-modal}, this direction is still under-explored. 
    Multi-modal learning can help in situations where the data of one mode are insufficient, and co-learning with other modes of data can compensate for such deficiency.
    A good practice would be to fuse visual data (e.g. images, videos of POIs, traffic, travel recordings), text data (e.g. user ratings, blogs, public announcements), and audio data into standard modeling processes, given that the general multi-modal techniques are relatively mature. However, the major bottleneck lies in data collection. Often, the multi-modal data important to our modeling aren't easy to collect because of privacy, cost, and other issues. Therefore, how to collect multi-modal data feasibly and then design new applications to meet more kinds of user needs is a promising trend.

    
    \item[$\bullet$] \textbf{Unified large models.}  Most route recommendation models only serve specific goals, and there is a need to study unified tasks and design very general large models. Recently, affected by the amazing progress made to the large models such as GPT \cite{floridi2020gpt} and Midjourney \cite{borji2022generated}, an interesting question would be how we can integrate them into route-designing processes such that pre-gained knowledge can be transferred and enhance current models. One example in the urban computing field is CityGPT~\cite{anonymous2024citygpt} proposed in WAIC 2023, which is a large model driven by public affairs and the joint services of multiple city departments. Up until now, incorporating large models for individual uses is still an open problem. For example, an interesting service would be to provide users with a personal travel manager who fully understands the user's needs in various situations (commuting, traveling, etc.) and can design suitable routes accordingly.
    If we particularly focus on large language models (LLMs), how to inject domain knowledge into LLMs, achieving semantic collaboration and cross-domain knowledge fusion, is an exciting direction for route recommendation. A latest survey~\cite{Lin2023HowCR} has reviewed many works that use LLMs to improve recommender systems, and similarly, the methodology could be used in route recommendation.

\end{itemize}

\section*{Acknowledgments}
This work was supported in part by the National Natural Science Foundation of China (Grant Nos. 62176221, 62276215, 61976247, 62302405), China Postdoctoral Science Foundation (Grant No. 2023M732914), Fundamental Research Funds for the Central Universities (Grant No. 2682023CX035), and the Natural Science Foundation of Sichuan Province (Grant No. 24NSFC2348).
\bibliographystyle{elsarticle-num}
\bibliography{Bibliography}

\end{document}